\documentclass[letterpaper]{article}

\usepackage[T1]{fontenc}

\usepackage{geometry}
\usepackage{setspace}
\usepackage{amsmath}
\usepackage{url}

\usepackage{graphicx}

\usepackage{booktabs}
\usepackage{multirow}

\usepackage{appendix}
\newcommand{\NotificationStartDate}{1st of January 2024}
\newcommand{\NotificationEndDate}{28th of February 2026}
\newcommand{\NotificationMonths}{26}
\newcommand{\NotificationPlumes}{2,776}
\newcommand{\NotificationSources}{458}
\newcommand{\NotificationCountries}{25}

\newcommand{\NotificationTones}{56,000}

\newcommand{\NumberOfSites}{3,800}
\newcommand{\NumberOfImagesMonthly}{50,000}

\def\scititle{
	Artificial intelligence for methane detection: from continuous monitoring to verified mitigation
}

\title{\bfseries \boldmath \scititle}

\usepackage{authblk}
\author[1]{Gonzalo~Mateo-Garcia$^{\ast\dagger}$}
\author[1,6]{Anna~Allen$^{\ast\dagger}$}
\author[1,4]{Itziar~Irakulis-Loitxate$^{\dagger}$}
\author[1]{Manuel~Montesino-San~Martin}
\author[1]{Marc~Watine}
\author[1]{Cynthia~Randles}
\author[1]{Tharwat~Mokalled}
\author[1]{Alma Raunak}
\author[1]{Carol Castañeda-Martinez}
\author[1]{Juan Enmmanuel Jonhson}
\author[4]{Javier~Gorroño}
\author[3]{James~Requeima}
\author[1]{Claudio~Cifarelli}
\author[4,5]{Luis~Guanter}
\author[2]{Richard~E.~Turner}
\author[1]{Manfredi~Caltagirone}

\affil[1]{International Methane Emissions Observatory, United Nations Environment Programme}
\affil[2]{Department of Engineering, University of Cambridge}
\affil[3]{Vector Institute, University of Toronto}
\affil[4]{Research Institute of Water and Environmental Engineering (IIAMA), Universitat Politècnica de València (UPV)}
\affil[5]{Environmental Defense Fund}
\affil[6]{Department of Computer Science and Technology, University of Cambridge}

\date{*To whom correspondence should be addressed; E-mail: anna.vaughan@un.org and gonzalo.mateogarcia@un.org\\$^{\dagger}$ These authors contributed equally to this work}

\begin{document} 

\maketitle

\begin{abstract} \bfseries \boldmath

 Methane is a potent greenhouse gas, responsible for roughly 30\% of warming since pre-industrial times. A small number of large point sources account for a disproportionate share of emissions, creating an opportunity for substantial reductions by targeting relatively few sites. Detection and attribution of large emissions at scale for notification to asset owners remains challenging. Here, we introduce MARS-S2L, a machine learning model that detects methane emissions in publicly available multispectral satellite imagery. Trained on a manually curated dataset of over 80,000 images, the model provides high-resolution detections every two days, enabling facility-level attribution and identifying 78\% of plumes with an 8\% false positive rate at 697 previously unseen sites. Deployed operationally, MARS-S2L has issued \NotificationPlumes{} notifications to stakeholders in \NotificationCountries{} countries, enabling verified, permanent mitigation of six persistent emitters, including a super-emitter in Algeria that had been releasing approximately 27,000 tonnes of methane annually for at least a decade and a previously unknown emitter first identified by MARS-S2L. These results demonstrate a scalable pathway from satellite detection to quantifiable methane mitigation.

\end{abstract}
\newpage
% \linenumbers

\noindent

\section{Introduction}

Methane is the second-largest contributor to human-induced global warming, responsible for
approximately 30\% of the warming observed since pre-industrial times \cite{stocker2014climate,dean2018methane}. Given its climatic effect is realised within a single decade, curbing methane emissions is an effective near-term intervention to moderate global temperature rise.\\

A large share of anthropogenic methane originates from a relatively small number of high-volume point sources. Recent inventories indicate that the largest 10\% of oil-and-gas facilities account for approximately 40\% of sectoral emissions \cite{lauvaux2022global}. One illustrative case is the Hassi Messaoud field in Algeria, where satellite images acquired in
2021 revealed persistent releases amounting to tens of kilotonnes of CH\(_4\)\,yr\(^{-1}\) that had gone unrecognised for years \cite{guanter2021mapping,sanchez2021mapping,irakulis2022satellite}. Currently, however, the global community lacks a systematic, timely way to discover and track large emitters before potentially decades of leakage accrue.\\

The development of real-time monitoring systems that detect emissions and trigger alerts for asset owners and policy makers remains an unrealised goal, but is now potentially becoming achievable. Expanding satellite coverage and advances in Earth observation instrumentation are now bringing this long-standing ambition within reach. Among the existing constellations, Sentinel-2 (S2) and Landsat provide public, global coverage imagery every two days \cite{li2020global} and are capable of detecting emissions greater than 1 t/hour ~\cite{ehret_global_2022,gorrono2023understanding}. In addition, their high spatial resolution  (20-30 meters) enables the attribution of emissions to specific facilities, a necessary step to enable mitigation action and drive accountability. As the spectral resolution of these instruments is relatively coarse, however, detecting the methane signal reliably in this very large volume of data poses a substantial technical challenge.\\

Early studies using these platforms focused on applying fixed threshold filters together with manual inspection to enable the visualisation and quantification of methane plumes \cite{varon2021high,irakulis2022satellites}. Given that there are millions of square kilometres of imagery over zones with high numbers of emitters, however, manually identifying plumes rapidly becomes an infeasible task. Machine-learning approaches are a natural solution to alleviate this problem, and have been applied to automate the detection of methane using other sensors \cite{schuit2023automated, balasus2023blended, joyce2022using,jongaramrungruang2019towards,ruuvzivcka2023semantic}. For multispectral data, however, existing efforts have either been confined to narrow geographic domains with homogeneous backgrounds \cite{vaughan2023ch4net,zortea_detection_2023,zhao2024data} or idealised studies using synthetic data \cite{joyce_using_2023,radman_s2metnet_2023,rouet2024automatic}.\\

Here we present \textbf{MARS-S2L} (the Methane Alert and Response System - Sentinel 2 and Landsat), the first large scale, operational methane emission monitoring model for Sentinel-2 and Landsat imagery. MARS-S2L is a machine learning model which automatically detects plumes in multispectral imagery. We first hand annotate a large dataset of plumes, and use this to train the model. We then evaluate MARS-S2L in a diverse range of regions globally, and detail the results of \NotificationMonths{} months of operational deployment and multiple cases of successful mitigation action based on MARS-S2L alerts.\\

\section{MARS-S2L}

\subsection{Dataset}\label{sec:dataset}

A major roadblock to developing a large scale machine learning model for this task has been the lack of a training and evaluation dataset of real methane plumes across a diverse range of regions globally. We therefore begin by building a large dataset of emission events, the MARS-S2L dataset. As large methane emission events are localised and relatively rare, this presents a significant challenge, with previous studies forced to examine only limited areas \cite{vaughan2023ch4net,zortea_detection_2023,zhao2024data}, rely on synthetic data \cite{joyce_using_2023,radman_s2metnet_2023,rouet2024automatic}, or data taken days before a ground truth plume detection \cite{rouet2024automatic}. To ensure the trustworthiness and suitability of a global detection model for operational deployment, it is necessary to have a large dataset of real plumes in multispectral imagery verified by experts. The MARS-S2L dataset utilises multispectral observations from Sentinel-2 and Landsat. We first compile a list of 1,315 sites by curating a list of potential emitters using third-party indicators of emissions, including TROPOMI methane hotspots, known infrastructure from databases such as Rystad Energy and the Oil and Gas Infrastructure Mapping (OGIM) dataset~\cite{ogim_2023}, detections from hyperspectral imagery, and prior reports from the scientific literature. To identify plumes, we hand label images at each site. This is a time consuming task, with hand labelling often taking up to five minutes per image. The resulting dataset contains 87,929 images spanning January 2018 to December 2024. These include 5,534 positive emission detections across 1,315 distinct emitter sites, representing a diverse set of regions globally which cover an estimated 70\% of upstream oil and gas production (Figure~\ref{fig:1}). For a detailed description of the dataset see materials and methods.
\\

\subsection{Model}

MARS-S2L is implemented as a simple, flexible UNet architecture. Inputs are six multispectral bands common to Sentinel-2 and Landsat for both the current pass and the most similar cloud free pass from the past 3 months, cloud masks generated using CloudSEN12 \cite{aybar_cloudsen12_2022} and northward and eastward 10m wind from ERA5-Land reanalysis \cite{munoz2021era5} or GEOS-FP for offshore platforms \cite{lucchesi2013file}. We also include multi-band multi-pass (MBMP)~\cite{irakulis2022satellites} differencing of the two multispectral passes as an auxiliary channel. The output is a probabilistic plume mask, which is converted to a scene-level probability by thresholding the per-pixel values (see materials and methods). As the number of plumes is small with respect to the total number of images, we additionally simulate plumes in images without plumes at training time following the end-to-end simulation methodology developed by Gorroño et al.~\cite{gorrono2023understanding} (see materials and methods). We utilise the same model on all geographical areas except for offshore where we additionally fine-tune the model to adapt to the significantly different spectra of water.

\subsection{Deployment and notification}\label{sec:systemdesign}

To be operationally effective, MARS-S2L detections must be accessible, interpretable, and actionable by analysts engaging with governments and operators. To facilitate this we develop the PlumeViewer, an interface that displays model outputs alongside historical predictions for monitored sites. Each day at 06:30 GMT, new Sentinel-2 and Landsat imagery over \NumberOfSites{} monitored emitters is automatically processed by MARS-S2L. New sites are constantly being added. If the detection is recent and can be attributed to a facility in the ground, a formal notification is issued to the government of the country and operator and MARS case-managers engage with the parties to begin working towards mitigation action. For the formal notification process, we prepare an official report for authorities where it is included a picture of the retrieved plume, the name, location, potential owner and operator of the identified facility as well as its quantified flux rate and the number of previous detection on the same facility. This report is submitted to a point of contact designated by the country authorities. Case managers follow up after few days to find out if mitigation action has been taken and requesting a feedback form with information about the event. \\

As there is no comprehensive, up-to-date global database of asset ownership and ownership frequently changes, identifying the responsible operator and engaging with them or the relevant government authority requires human oversight. This precludes full automation of the notification component of the monitoring pipeline. Consequently, all predictions generated by the system are manually reviewed prior to notification. Within these operational constraints, the MARS-S2L model plays a key role as a filtering mechanism: its principal function is to reduce the overwhelming volume of daily satellite imagery to a manageable subset of high-probability emission candidates for analyst verification. In this context, we prioritize high recall to ensure that the majority of plumes are detected, while also maintaining a low false positive rate (FPR). \\

 \begin{figure}
     \centering
     \includegraphics[width=1\linewidth]{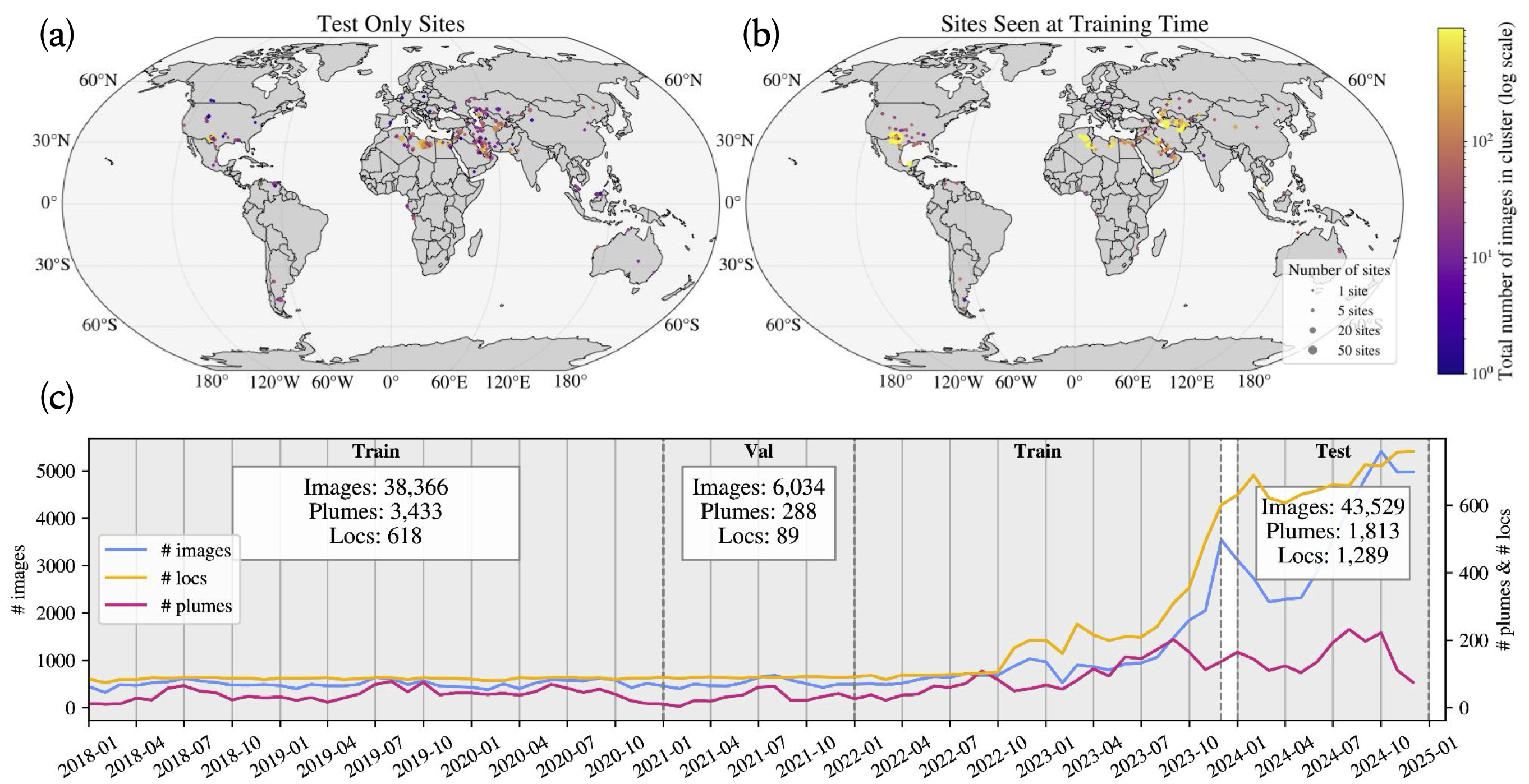}\\
     \caption{\textit{\textbf{Characteristics of the MARS-S2L dataset} We compile a large dataset of emissions for training and evaluating machine learning models. In total, 87,929 images containing 5,534 emissions over 1,315 distinct emitters globally are included in the dataset. \textbf{Top} emitter locations and for the test (a) and training (b) sets. \textbf{Bottom} Time series of monthly number of images, plumes, and distinct locations. Shaded areas indicate the split of the dataset in train, validation, and test subsets.}}
     \label{fig:1}
\end{figure}

\section{MARS-S2L performance}

We begin by evaluating model performance on a held-out test set consisting of all images from 2024, a total of 43,529 images with 1,813 manually verified plumes across 1,289 different emitters. We compare MARS-S2L against two established baselines: multiband-multipass (MBMP) differencing and CH4Net. MBMP differencing \cite{varon2021high,gorrono2023understanding} consists of thresholding the original physics based method used for retrieving methane plumes in multispectral imagery, while CH4Net \cite{vaughan2024ch4net} represents the most extensive machine learning approach currently available for this task. Full implementation details for both baselines are provided in Methodology. Figure 2 (bottom; global results) presents a comparison of recall and false positive rate (FPR) across different emission intensities. MARS-S2L achieves a recall of 0.79 and a false positive rate of 0.08. This represents a substantial improvement over the baselines: MBMP achieves similar recall (0.79) but with an elevated FPR of 0.69, while CH4Net reduces the FPR to 0.10 but suffers from low recall (0.47). Figure 2 shows the number of images required to be reviewed to identify a certain number of plumes where images are ordered by the predicted probability of containing an emission. This demonstrates that an analyst working with MARS-S2L needs to review far fewer images than one working manually to identify the same number of plumes. For example, identifying 500 plumes manually would require the analyst to search through 22 times as many images than using MARS-S2L. In practical terms this unlocks the ability of a small team of 5 analysts to handle incoming data and create alerts, as opposed to requiring dozens which would be financially unfeasible.  \\

 \begin{figure}
     \centering
      \includegraphics[width=0.9\linewidth]{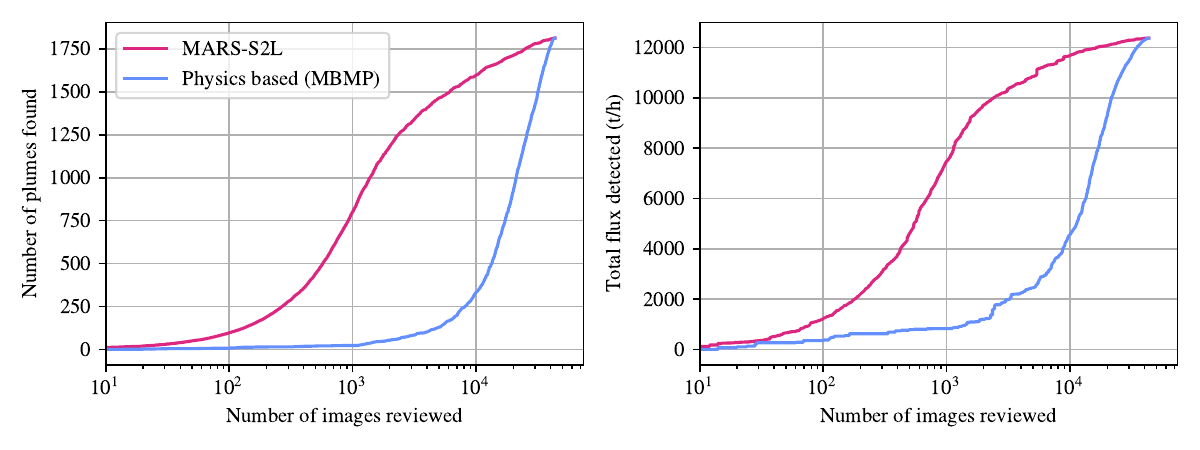}\\
     \includegraphics[width=1\linewidth]{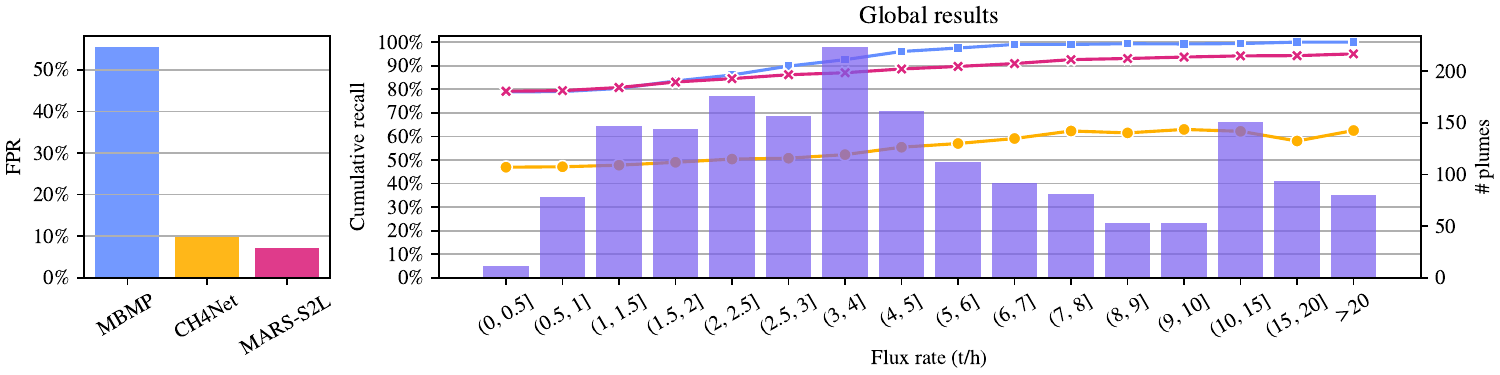}
     \caption{\textit{\textbf{MARS-S2L enables large scale processing of Sentinel-2 and Landsat images.} Top: Images required to be reviewed to identify different amount of plumes when ordered by model probability. This indicates that incorporating AI enables analysts to review around 22 times fewer images, making the monitoring feasible to be tackled by a small team. Bottom: False positive rate and cumulative recall of the MARS-S2L model compared to CH4Net and MBMP.}}
     \label{fig:draft}
\end{figure}

A key question for deployment is the model’s ability to generalize to locations not represented in the training set. We assess this generalization capability in two ways. First, we evaluate performance at sites excluded from training, a total of 697 of the 1,289 test sites. At these held-out locations, MARS-S2L achieves a recall of 0.78 and a false positive rate (FPR) of 0.08, nearly identical to its performance at sites for which training data are available (recall 0.79, FPR 0.07) indicating robust generalization. Second, we evaluate how MARS-S2L performs when scanning arbitrary scenes by quantifying the FPR across diverse background types. To this end, we apply MARS-S2L to the full CloudSEN12 dataset \cite{aybar_cloudsen12_2022,aybar_cloudsen12_2024}, a large-scale collection of representative terrestrial imagery (see Supplementary Information). Across 10,434 scenes, the model yields a false positive rate of 0.09. We note that these figures are reported with a scene probability threshold of 0.5. If a lower false positive rate is required, different probability thresholds can be used. For example, setting a threshold of 0.98 results in a false positive rate of 0.01 with a recall of 0.58. \\

To evaluate the sensitivity of MARS-S2L on emissions with a known flux rate, we apply the model to all imagery from the Stanford controlled release experiments conducted in 2021 and 2022 \cite{sherwin_single-blind_2023, sherwin_single-blind_2024}, in which methane was released at known rates during Sentinel-2 and Landsat overpasses. These locations were all unseen at training time. For the 2021 experiment, we analyzed eight scenes (six Sentinel-2 and two Landsat), excluding one Sentinel-2 image due to cloud cover, resulting in seven positive and one no emission sample with release rates up to 7.5 t/h. The 2022 experiment posed a more challenging test, comprising ten positive and five no emission samples with emissions between 0.75 and 1.5 t/h. As shown in Figure~\ref{fig:controlled_releases} (left), MARS-S2L successfully detected all emission events greater than 3 t/h and did not produce any false positives. For emissions of 0.75–2 t/h, the model identified five positives out of thirteen. Performance was slightly lower on Landsat imagery, likely due to its reduced spatial resolution and under-representation in the training data (see Table~\ref{tab:splits}).\\

\section{Regional performance}
We next evaluate model performance across 12 geographically distinct regions encompassing major oil and gas extraction areas. These regions vary in both the number of satellite samples and the prevalence of manually verified methane plumes. The United States, Turkmenistan, Algeria, Libya, and the Arabian Peninsula represent areas with both high sample density and a large number of confirmed emissions. In contrast, Uzbekistan and Kazakhstan, Iran, Egypt, offshore sites, Iraq, and Syria have similarly high sample counts but fewer than 100 identified plumes per region. Venezuela, meanwhile, constitutes a lower-data setting, with comparatively fewer samples and detections. \\

\begin{figure}
     \centering
     \includegraphics[width=1.0\linewidth]{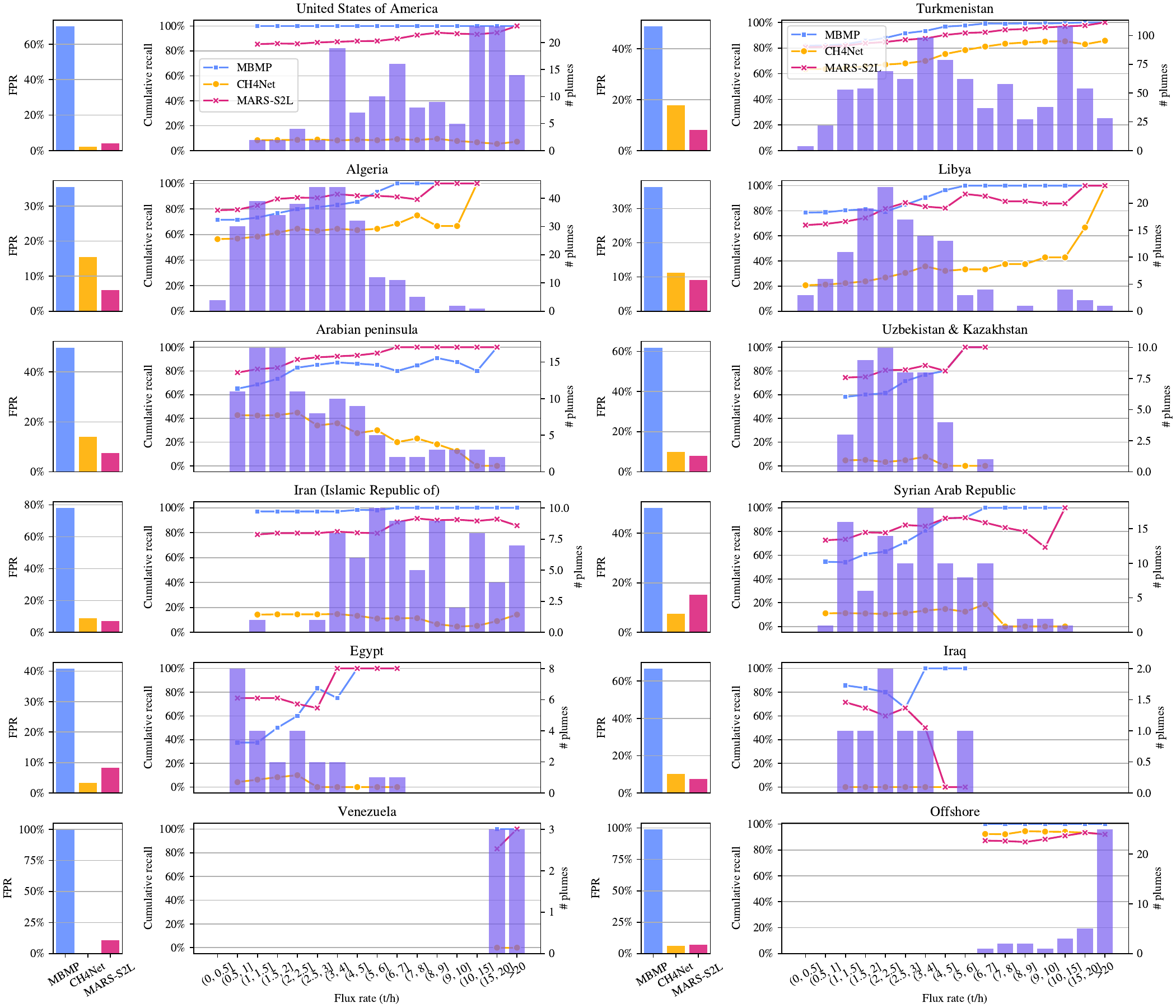}\\
     \caption{\textit{\textbf{Multi-regional evaluation of model performance.} False positive rate (FPR) and cumulative recall of the MARS-S2L model compared with multi-band multi-pass (MBMP) thresholding and CH4Net.
     Results are stratified by flux rate over the 12 selected geographical areas. 
     MARS-S2L provides skilful identification of emissions showing a false positive rate under 10\% and recall around 80\% for all plumes and up to 90\% for plumes of more than 5 t/h.}}
     \label{fig:2}
\end{figure}

Results in Figure~\ref{fig:2} present the cumulative recall and FPR for MARS-S2L across the 12 geographic regions, compared against the CH4Net and MBMP baselines. MARS-S2L demonstrates consistent performance across most areas, with recall above 0.80 for all plume strengths in the United States, Turkmenistan, and Algeria. The FPR remains below 0.10 in the majority of regions, with the exceptions of Syria (0.153) and Venezuela (0.11). For emissions $>$ 5 t/h, recall reaches 0.90 or higher in all regions with large numbers of plumes, including Turkmenistan, the Arabian Peninsula, Libya, Algeria, and the United States. Detection performance could be further optimized by tuning probability thresholds on a per-site or per-region basis. Overall, the analysis indicates that MARS-S2L delivers high recall and low false positive rates across a wide range of emission scenarios and geographic contexts.

\section{Operational deployment results}

We next deploy MARS-S2L and generate formal notifications for governments and corporate end users. The model is deployed as part of the MARS system at IMEO, automating the detection of methane emissions at over \NumberOfSites{} sites globally processing monthly \NumberOfImagesMonthly{} images on average at the time of this analysis. We note that the model is frequently retrained on all available data, and hence is not the exact model for which results are reported above where data are held out for analysis. \\

\begin{figure}
     \centering
    \includegraphics[width=1.02\linewidth]{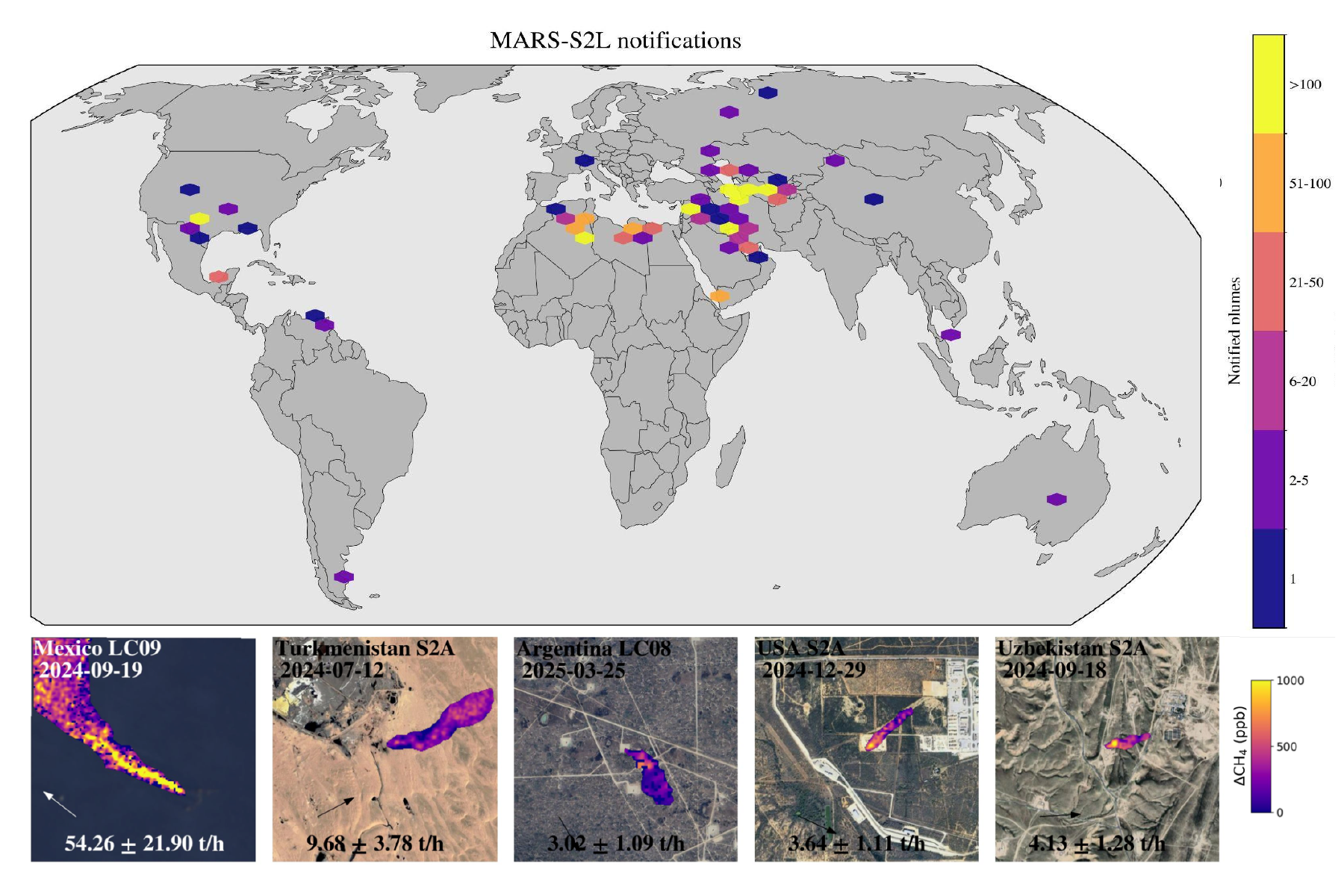}
    \caption{\textit{\textbf{Model notifications during 26 months of operational deployment.} (a) Map with MARS-S2L notified detections in red. During this period, MARS-S2L detected \NotificationPlumes{} emissions across \NotificationCountries{} countries that triggered a formal notification to governments and operators. Examples of some of these notifications are shown for events in Mexico, Turkmenistan, Argentina, the US and Uzbekistan (b). All plume detections are validated by IMEO analysts prior to notification.}}
     \label{fig:3}
\end{figure}

We report results for a 26-month period from the \NotificationStartDate{} to the \NotificationEndDate{}. During this phase, the model produced \NotificationPlumes{} detections from \NotificationSources{} unique oil and gas production sources in \NotificationCountries{} different countries that resulted in government and corporate stakeholder notifications. Assuming a 3 hour duration of each emission, a very conservative estimate since many of the notified sources are persistent, approximately \NotificationTones{} tonnes of methane were released into the atmosphere by all these events. The locations of all detections and notifications together with examples of multiple notified plumes are shown in Figure~\ref{fig:3}. For a detailed description of the operational detection process, see materials and methods.\\

\section{Successful mitigation cases}\label{sec:successful_mitigation_cases}

Notification of emissions is only the first step toward eliminating them. Effective mitigation requires corrective action to resolve the underlying issue, followed by regular satellite monitoring to ensure that emissions do not recur. The MARS-S2L system contributes to this process through the detection of plumes, the issuance of notifications to relevant stakeholders, and verification via negative detections during post-mitigation monitoring.\\

During the first 20 months of operational deployment, six successful mitigation cases of persistent methane emitters were recorded across Algeria, Libya, Kazakhstan, Yemen, Argentina, and Turkmenistan~\cite{unep_imeo_2024,unep_imeo_2025}. Among these, two cases stand out: the elimination of a methane emitter that had been active for at least 25 years in Algeria, and the first mitigation of a new site discovered by the MARS-S2L model in Libya. The remaining cases are described in detail in materials and methods.\\

At Hassi Messaoud, Algeria, emissions originated from a gas disposal facility that had been persistently active since at least 1999 ~\cite{guanter2021mapping,sanchez2021mapping,irakulis2022satellite,unep2024eye}, making it one of the oldest persistent emitters on record~\cite{unep2024eye}. From June 2024 onwards, MARS-S2L detections were used to make multiple notifications to relevant parties on the leak. Over this time, IMEO engaged further with relevant authorities in Algeria, including the designation of focal points to respond to notifications and hosting in-person training in Algeria where this case was presented again~\cite{unep_imeo_2024}. The focal points committed to investigating the case, and on October 14th emissions ceased. Detailed feedback from national and operator focal points via MARS response forms and public statements made by operators during COP29~\cite{cop29_imeo_session_2024} confirmed that this mitigation action was taken in response to MARS notifications. The mitigation prevented an estimated 27,500 tonnes of methane per year, equivalent over a 20-year horizon to the annual CO$_2$ emissions of approximately 500,000 standard passenger vehicles (see Section~\ref{rwi} for calculation details). Figure 5 (a) shows a time series of emissions observed for this site in 2024, with examples of detections in Figure 5 (b).\\

In Libya, MARS-S2L was run on the locations of over 300 oil and gas facilities. This led to the identification of new emitters, including an upstream production site in the Faregh–Argub basin (Figure S13). A notification was issued, prompting investigation by the operator, who determined that emissions originated from mud tanks storing oil-based drilling fluids. The company subsequently reused the stored mud in operations, eliminating the emission source~\cite{unep_imeo_2025}. This case highlights the growing integration of MARS-S2L into early-stage mitigation workflows where the model detects previously unknown emitters and issues notifications to trigger mitigation action.\\

\begin{figure}
     \centering
     \includegraphics[width=1\linewidth]{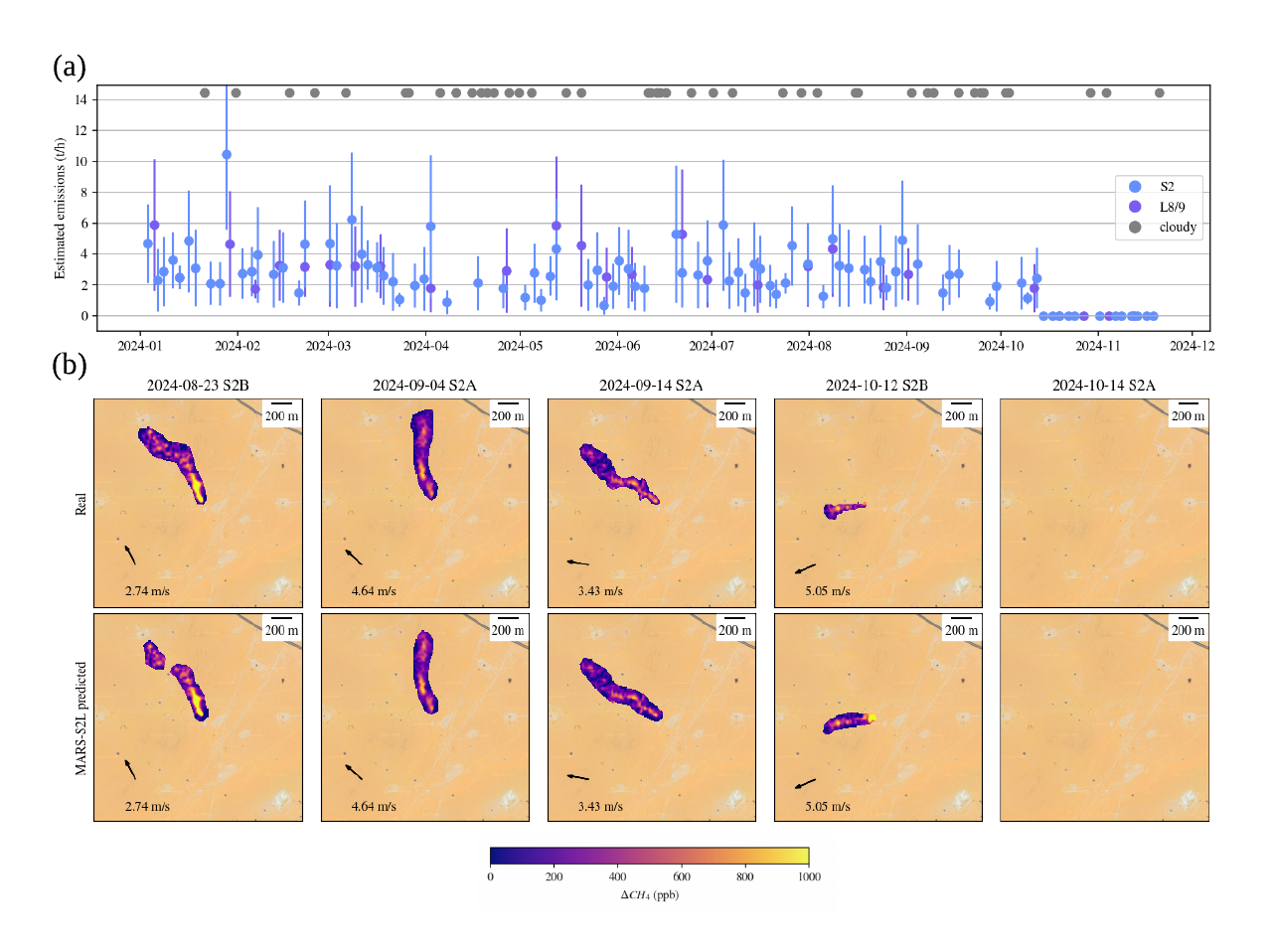}
     \label{fig:figure5}
     \vspace*{-1.5cm}
     \caption{\textit{\textbf{MARS-S2L predictions for a mitigated emitter in Hassi Messaoud, Algeria} A timeseries of emissions with quantification is shown in (a), showing the mitigation of this source on the 14th of October 2024. Examples of MARS-S2L predictions compared to expert-annotated ground-truth are shown in (b). MARS-S2L successfully detects emissions from this source until its cessation.}}
\end{figure}

\section{Discussion}

We have introduced MARS-S2L, the first machine learning system to detect real-world methane emissions on a large scale. Trained on a comprehensive, expert-validated dataset, MARS-S2L achieves robust performance across diverse geographic regions, with a recall of 0.79 and a false positive rate of 0.07. Over an initial 26-month operational period, the system enabled the near-real-time detection of \NotificationPlumes{} emission events which have already been translated into real world action mitigating multiple sources.\\

These results demonstrate that large-scale, operational methane emission detection is not only technically feasible but already delivering real-world impact. The successful mitigations of persistent emitters across different regions provide a clear example of how an operational system can be used to submit stakeholder notifications and trigger real-world action. With over three thousand emitters now under continuous observation with Sentinel-2 and Landsat, supported by ongoing collaboration between IMEO, national governments, and industry stakeholders, MARS-S2L holds significant potential for mitigating methane emissions at scale. In addition to notifying stakeholders, validated plume detections are published bi-weekly on IMEO’s Eye on Methane public data portal (\texttt{methanedata.unep.org}), where they are accessed by a broad cross-section of society, including energy companies, policymakers, researchers, NGOs, media, and private sector organizations. \\

Improvements to the system are possible along a number of axes. MARS-S2L is expanding to leverage hyperspectral sensors, including the PRecursore IperSpettrale della Missione Applicativa (PRISMA), the Environmental Mapping and Analysis Program (EnMAP) and the Earth Surface Mineral Dust Source Investigation instrument (EMIT)~\cite{ruzicka_operational_2026}. Additionally a similar system can tackle detections in emerging platforms within the growing methane point-source monitoring ecosystem, such as Sentinel-3 \cite{pandey2023daily} and the Visible Infrared Imaging Radiometer Suite \cite{de2025daily}
 These additions will expand spatial and temporal coverage, bringing the system closer to continuous, near-real-time global monitoring. Advances in remote sensing and machine learning, such as large-scale foundation models~\cite{hong2024spectralgpt,cha2023billion}, simulation-based data augmentation, and the incorporation of diverse, high-quality training datasets offer promising directions for further improving predictive skill. Moreover, the continuously growing dataset of validated detections provided by IMEO will further support ongoing improvements in model robustness and operational effectiveness. \\

The potential of this approach extends far beyond methane. Large emissions of other greenhouse gases, for example $CO_2$ and $NO_2$, can also be detected via remote sensing, and are equally suited to automated monitoring. By combining similar machine learning systems with increasingly comprehensive asset databases and active engagement from governments and operators, a path is emerging toward a future of global, real-time greenhouse gas emission surveillance. In such a system, high-emission events could trigger instant alerts to responsible parties anywhere on Earth, enabling timely, targeted mitigation at unprecedented scale. Realizing this vision would be a substantial step toward accountable, data-driven climate action.

\section*{Acknowledgments}
We dedicate this paper to the memory of our coauthor and friend, Manfredi Caltagirone. As the founder and head of IMEO, his vision, dedication, and leadership enabled this work and left lasting impacts throughout the field of methane emissions reduction.
We thank Meghan Demeter, Giulia Bonazzi, Florencia Carreras and Konstantin Kosumov as the MARS members in charge of country engagement and dealing with the notification process, as well as the rest of the IMEO group, including the OGMP 2.0 group, and the EDF for their support, advice, and fruitful discussions. We thank the MethaneSAT developer team: Chris Hairfield, Susan Kasper, Carlos Cervantes, Dave Calhoun, Courney Maimon, Ravi Savajiyani and Cori Mortimer for their support as co-developers of the PlumeViewer application. 
We thank the members of the IMEO Science Oversight Committee for their feedback. We thank Luis Gomez-Chova, Cesar Aybar and Enrique Portales-Julia from the Universitat de València (UV) for adapting the CloudSEN12 cloud detection model to Landsat and for the support of the georeader package; and the Land and Atmosphere Remote Sensing (LARS) team at the Universitat Politècnica de València (UPV) for their support of this work supported by grants PID2023-148485OB-C21 and C22 funded by MCIU/AEI/10.13039/501100011033 ERDF, EU. G.M.-G. and A.A. gratefully acknowledge the Frontier Development Lab (fdleurope.org), where the collaboration among the main authors of this paper was initiated.\\

\paragraph*{Author contributions:}
M.C, A.A, C.C and I.I-L conceptualised the project. A.A collected and labelled the training dataset, designed the AI model and developed the initial version. G.M-G developed the PlumeViewer and operational deployment pipeline assisted by J.J. A.A and G.M-G developed the final version of the AI model. I.I-L led the identification of new sites and development of the evaluation dataset and notification process, assisted by M.M.S-M, M.W, G.M-G, A.A., A.R. and C.C.-M. R.E.T and J.R provided assistance with machine learning techniques. J.G provided assistance with the physics-based simulation scheme. T.M led the notification in the mitigation case study. A.A and G.M-G wrote the initial draft of the manuscript. L.G, C.R, and all aforementioned authors provided feedback on results throughout the project and contributed to the final version of the manuscript. 

\paragraph*{Competing interests:}
There are no competing interests to declare.

\paragraph*{Data and materials availability:}
Model weights, and data are publicly available under a Creative Commons license CC BY-NC-SA 4.0 DEED (Attribution-NonCommercial-ShareAlike 4.0 International) on HuggingFace at \texttt{https://huggingface.co/datasets/UNEP-IMEO/MARS-S2L}.
The code for training, evaluation and reproducing the figures is publicly available at \texttt{https://github.com/UNEP-IMEO-MARS/marss2l}.

\clearpage

\bibliography{science_template}
\bibliographystyle{sciencemag}

\newpage

\renewcommand{\thefigure}{S\arabic{figure}}
\renewcommand{\thetable}{S\arabic{table}}
\renewcommand{\theequation}{S\arabic{equation}}
\renewcommand{\thepage}{S\arabic{page}}
\setcounter{figure}{0}
\setcounter{table}{0}
\setcounter{equation}{0}
\setcounter{page}{1}

\begin{center}
\section*{Supplementary Materials for\\ \scititle}

\author{
Gonzalo~Mateo-Garcia$^{1\ast\dagger}$, 
Anna~Allen$^{1,6\ast\dagger}$, 
Itziar~Irakulis-Loitxate$^{1,4\dagger}$, 
\\ Manuel~Montesino-San~Martin$^{1}$,  
Marc~Watine$^{1}$, 
Cynthia~Randles$^{1}$,\\
Tharwat Mokalled$^{1}$,
Alma Raunak$^{1}$,
Carol Castañeda-Martinez$^{1}$,\\
Juan Enmmanuel Jonhson$^{1}$,
Javier~Gorroño$^{4}$, 
James~Requeima$^{3}$, \\ 
Claudio~Cifarelli$^{1}$
Luis Guanter$^{4,5}$, 
Richard~E.~Turner$^{2}$, \\
and
Manfredi~Caltagirone$^{1}$\\
\small{\textsuperscript{1}International Methane Emissions Observatory, United Nations Environment Programme}\\
\small{\textsuperscript{2}Department of Engineering, University of Cambridge}\\
\small{\textsuperscript{3}Vector Institute, University of Toronto}\\
\small{\textsuperscript{4}Research Institute of Water and Environmental Engineering (IIAMA)},\\ \small{Universitat Politècnica de València (UPV)}\\
\small{\textsuperscript{5}Environmental Defense Fund},\\ 
\small{\textsuperscript{6}Department of Computer Science and Technology, University of Cambridge},\\
\small{\textsuperscript{*}To whom correspondence should be addressed; E-mail:  anna.vaughan@un.org} and \small{gonzalo.mateogarcia@un.org}\\
\small{\textsuperscript{†} These authors contributed equally to this work}}
\end{center}

\subsubsection*{This PDF file includes:}
Materials and Methods\\
Supplementary Text\\
Figures S1 to S27\\
Tables S1 to S10\\

\newpage

\subsection*{Materials and Methods}

Throughout the next sections we use a threshold of 0.5 for all predictions unless otherwise stated.

\label{sec:methods}
\subsubsection{Input datasets}

MARS-S2L utilises data from five earth observation satellites. Sentinel-2A (launched 2015), Sentinel-2B (launched 2017), and Sentinel-2C (launched 2024) are Copernicus program satellites operated by the European Space Agency, capturing high-resolution images in 13 spectral bands using the Multispectral Imager (MSI) instrument. Landsat 8 (launched 2013) and Landsat 9 (launched 2021) are part of the National Aeronautics and Space Administration (NASA) and the United States Geological Survey (USGS) Landsat program and utilize the Operational Land Imager (OLI), providing multispectral images with nine spectral bands. \\

MARS-S2L is trained using the following common bands of OLI and MSI: blue (MSI: ~490 nm, OLI: ~482 nm), green (MSI: ~560 nm, OLI: ~561 nm), red (MSI: ~665 nm, OLI: ~655 nm),  near infrared (MSI: ~842 nm, OLI: ~865 nm) and the two short wave infrared (SWIR) bands at wavelengths ~1610 nm (SWIR1) and ~2190 nm (SWIR2). We exclude aerosol and cirrus bands of both sensors as they have significantly different spatial resolution.  Bands of both sensors with spatial resolution higher than 10 m are interpolated to 10 m using bi-cubic interpolation. Although there are slight differences in central wavelength, band width and resolution between both sensors, these differences are not sufficient to impact model performance. Several auxiliary data channels are provided. For each image cloud and cloud shadow masks were computed using the CloudSEN12 model \cite{aybar_cloudsen12_2022}. Wind vectors are obtained from the ERA5-Land reanalysis \cite{munoz2021era5} or NASA GEOS-FP \cite{lucchesi2013file} for offshore locations.\\

\subsubsection{Image processing, labelling and quantification}

The first challenge in developing the dataset is compiling a comprehensive list of emitter sites. We begin by manually curating a list of potential emitters using third-party indicators of emissions, including TROPOMI methane hotspots, detections from hyperspectral imagery, known infrastructure from databases such as Rystad Energy and the Oil and Gas Infrastructure Mapping (OGIM) dataset, and prior reports from the scientific literature. \\

For each of these locations Sentinel-2 and Landsat images are downloaded for a 2$\times$2 km$^2$ square around the source. Each image is processed with CloudSEN12 and discarded if it contains more than 50\% cloud, cloud shadow, or missing pixels. For clear images, the retrieval is computed using the Multi-Band Multi-Pass (MBMP) ratio \cite{irakulis2022satellites}. We select the reference image as the most similar cloud-free image measured in the visible and SWIR1 band and restrict to images acquired over the last 4 months. For offshore platforms, we use the Multi-Band Single-Pass ratio (MBSP)~\cite{irakulis-loitxate_satellites_2022}. \\

From the retrieval image, a binary plume mask is manually annotated to delineate the plume using the Computer Vision Annotation Tool~\cite{CVAT_2023} and labelling functionality within PlumeViewer (see Appendix A for a full description of the PlumeViewer tool). This annotation process is inherently challenging and time-consuming (requiring several minutes per image) primarily due to retrieval artifacts in the MBMP method. While MBMP theoretically measures methane transmittance under ideal conditions, violations of its assumptions—such as surface reflectance changes between reference and target images, or atmospheric disturbances (e.g., cloud shadows, thin cirrus or smoke)—produce artifacts that mimic true methane signatures. Analysts must therefore carefully distinguish genuine plumes from artifacts by examining correlation with surface features or atmospheric phenomena in reference imagery; using also supporting evidence like wind direction and speed. 
Annotation is further complicated by the diffuse nature of plume boundaries or missing pixels near the swath edges. \\

All annotations were performed by a team of seven analysts. For positive examples we estimate the per-pixel concentration enhancement of methane ($\Delta$CH$_4$) and the flux rate in tonnes per hour (t/h) following Gorroño et al.~\cite{gorrono2023understanding}. Each validated detection in the test dataset has been inspected by at least 2 analysts. For plumes that are notified to governments and stakeholders a third confirmation is carried out by a different analyst.\\

The resulting dataset contains 87,929 images spanning January 2018 to December 2024. These include 5,534 positive emission detections across 1,315 distinct emitter sites (Figure~\ref{fig:1}). For model training and evaluation we split the dataset temporally into train (2018-2021 and 2022-2023; 3,433 plumes in 618 locations), validation (2021; 288 plumes in 89 locations) and test (2024; 1,813 plumes in 1,289 locations). Test data lie across 12 geographical areas covering major oil and gas extraction fields mostly in arid and semi-arid areas~(Figure~\ref{fig:1}). We note that while holding out such a large test dataset allows for a robust evaluation of model performance, large improvements are possible by adding this data to the model training prior to operational deployment. \\

\subsubsection{Test dataset coverage}\label{sec:test_dataset_coverage}

We hold out a large dataset for testing, covering all images gathered in the year 2024: 43,529 images with 1,813 manually verified plumes across 1,289 different sources. Images in our dataset are located in most of the major oil and gas extraction fields in arid and semi-arid regions, offshore, and Venezuela. \\

A substantial portion of emissions and sites are located in arid and semi-arid regions. This reflects both the geographical distribution of oil and gas infrastructure and the fact that plume detection in multispectral imagery is more feasible over uniform, high-albedo backgrounds and in cloud-free conditions. \\

Although the dataset includes some images over other oil- and gas-producing countries such as Russia, China, Australia, and Argentina, the number of verified emissions from these regions remains limited. Several factors contribute to the scarcity of detections, including persistent cloud cover that obstructs satellite observations, snow cover, frequent surface changes, low surface albedo, and the possibility that emissions fall below the detection thresholds of current satellite instruments. These conditions also hinder the identification of suitable observation pairs for MBMP retrieval.\\

Figure~\ref{fig:stats_marss2l_test} presents descriptive statistics of the MARS-S2L test dataset, disaggregated by region. The boxplot on the right illustrates that most validated plumes in arid and semi-arid regions fall within the range of 1 to 10~t/h. The left panel shows the standard deviation of the MBMP retrieval across regions, which serves as a proxy for retrieval noise. This noise level is relatively consistent across arid and semi-arid regions but is markedly higher in the offshore and Venezuela case studies. This elevated noise helps explain why only the largest plumes are detected in these areas. The central panel displays the mean top-of-atmosphere reflectance in the SWIR-2 band for detected plumes. As expected, arid and semi-arid regions exhibit high reflectance values (above 0.25), whereas the offshore and Venezuela case studies show significantly lower reflectance (below 0.05 offshore and approximately 0.15 in Venezuela). In such low-radiance conditions, the reflectance signal is inherently noisier, further limiting the ability to detect smaller plumes.\\

\subsubsection{Detection limit on Sentinel-2 and Landsat}\label{sec:mdl}

The emission frequencies recorded in our database can be described both by the actual magnitude of the emissions and by the ability of the monitoring system to detect them. Although true global emission frequencies follow a power law distribution ~\cite{lauvaux2022global}, the observed frequencies only match this pattern at higher emission rates. At lower rates, the observed frequencies diverge from the power law due to the detection limits of the monitoring system ~\cite{ehret_global_2022}. This is frequently modelled as a logistic function representing the Probability of Detection (PoD) curve. We fitted this model to our manually verified dataset of emissions in the year 2024 (Section~\ref{sec:dataset}) using least squares to estimate the probability of detection of this monitoring system. 
Following this analysis, the minimum flux rate detected in Sentinel-2 and Landsat is 2.46 and 2.18 tonnes per hour (t/h) respectively with a probability of detection of 90\%. This estimation is based on our validated test dataset. We note, however, that this varies significantly across extraction basins (Figure~\ref{fig:mdl_PoD}).

\subsubsection{Model design and architecture}\label{sec:class_score}

MARS-S2L is implemented as a UNet architecture \cite{ronneberger2015u}, with a total of 16 input channels.
 The backbone decoder is implemented as a simple and flexible UNet architecture \cite{ronneberger2015u} consisting of four encoder blocks (2D convolution layer, batch norm, ReLU activation, max pool) followed by four decoder blocks (transposed 2D convolution layer, 2D convolution layer, batch norm, ReLU activation, 2D convolution layer, batch norm ReLU activation) with skip connections between blocks of corresponding scale. Kernel sizes are 3 for all convolution layers and 2 for the max pooling layers. An schematic of the network is shown in Figure~\ref{fig:modelandplumeviewer}. MARS-S2L is trained with a pixelwise binary cross entropy loss with positive pixels upweighted by the per-pixel methane concentration $\Delta$CH$_4$ as in Ruzicka et al.~\cite{ruuvzivcka2023semantic}.\\

The output of the network is a per-pixel probability map. In order to provide the estimated quantification of the emission, we threshold this image by a predicted pixel probability of 0.5 to produce a binary mask, and multiply the output by the per-pixel methane enhancement ($\Delta$CH$_4$) and calculate the integrated mass enhancement (IME)~\cite{gorrono2023understanding} (figure~\ref{fig:controlled_releases}).\\

We opted to report scene-level metrics rather than segmentation metrics as they are more meaningful for the operational use of the model (Section~\ref{sec:systemdesign}). We calculate the scene-level score as the minimum threshold of the probability map such that there are at least 100 connected pixels in the output mask.  We choose the 100 threshold based on the size of the plumes in the dataset (Figure~\ref{fig:areaplume}). We observe only minor differences in the results when choosing other thresholds (Table~\ref{tab:scenelevelmetrics}).

\subsubsection{Model training procedure}\label{sec:model_training}

Training is performed using stratified sampling and a novel physics-based simulation scheme (Figure ~\ref{fig:simulation_images}) designed to compensate for the dataset imbalance of plumes and images across locations. During batch construction, we sample locations and binary indicators to select images with or without a plume. As it is desirable to train on real images where possible, plumes are simulated depending on the number of real plumes available at each location. For locations with no real plumes, we randomly sample an image and simulate a plume following the procedure below. In the case of one to five real plumes, a synthetic image is created with probability 0.9 and a real image is used otherwise. Finally, in cases where a site has more than five images with real plumes available, a synthetic image is created with probability 0.1 and a real image used otherwise. Based on the current distribution of the MARS-S2L training dataset, on average a plume is simulated approximately 50\% of the time. Although these thresholds are arbitrarily selected, changing these probability thresholds was found to have little impact on model performance.\\

To simulate a plume in an image, we use a physics based procedure based on the work of \cite{gorrono2023understanding}. This method simulates the per pixel transmittance of the plume in methane-absorbing bands of Sentinel-2 and Landsat (B11 and B12 for Sentinel-2 and B6 and B7 for Landsat). The input for this process is a real methane concentration image ($\Delta$CH$_4$) sampled from the positive images in our training dataset. This contrasts with other works that use synthetic plumes that may not produce realistic methane concentrations ~\cite{zortea_detection_2023,rouet2024automatic}. Specifically, for a plume from the training dataset, we crop the methane concentration image ($\Delta$CH$_4$) with the plume mask. With the $\Delta$CH$_4$ cropped image and the viewing geometry and solar angles of the clear scene (${\neg plume}$), we estimate the transmittance for bands B11 and B12 ($\tau_{B11}$, $\tau_{B12}$) (bands 6 and 7 of Landsat). The transmittance estimation is based on the MODerate resolution atmospheric TRANsmission (MODTRAN) radiative transfer model~\cite{MODTRAN}. We use MODTRAN to generate a look-up table (LUT) with the relationship between the methane enhancement ($\Delta$CH$_4$) and the transmittance at a fine spectral resolution $T_{\Delta\text{CH}_4}(\lambda)$ for different viewing geometries assuming a constant background concentration of 1800 ppb. For a given $\Delta$CH$_4$ and solar and viewing zenith angles, we use bicubic-spline interpolation of the LUT to obtain $T_{\Delta\text{CH}_4}(\lambda)$ which we then integrate over the spectral response function (SRF) of the satellite. The plume is then injected on the bands of the image following the equation~\cite{gorrono2023understanding}: 

\begin{align}
 BA &=  BA_{\neg plume} \frac{ \int E_g(\lambda) T_{atm}(\lambda) T_{\Delta \text{CH}_4}(\lambda) \text{srf}_{BA}(\lambda) d\lambda}{\int E_g(\lambda) T_{atm}(\lambda) \text{srf}_{BA}(\lambda) d\lambda}\\
   &= BA_{\neg plume} \cdot \tau_{BA}
\end{align}

Where $BA$ refers to band B11 or B12 of Sentinel-2 or B6 or B7 of Landsat, $BA_{\neg plume}$ is the pixel value of the plume-free image in the band, $T_{\Delta \text{CH}_4}(\lambda)$ is the transmittance for a concentration of $\Delta \text{CH}_4$ over the background, $E_g(\lambda)$ is the total solar irradiance, $T_{atm}(\lambda)$ is the atmospheric transmittance and the integral is taken over the SRF of the satellite for the band. We use a standard atmosphere simulation from LibRADTRAN~\cite{libradtran} for $E_g$ and $T_{atm}$ and assume constant surface reflectance over band wavelengths. \\

Wind plays a major role in detecting methane plumes. Higher winds disperse the plume more rapidly, making it difficult to detect. In addition, wind direction changes provide evidence to distinguish weak plumes from static artifacts. As weaker plumes are not visible at high wind speed, the simulation process takes into account the wind conditions of the clear image. Specifically, for a given image, the difference in wind speed between the clear and the sampled plume must be less than 1.5 m/s; in addition, if the wind speed of the clear image is higher than 9 m/s we do not simulate the plume. After sampling the plume, we rotate the methane concentration image to align it with the wind direction of the image to simulate into. We note that this \emph{wind consistent} procedure is required because the proposed detection model uses the wind field as input. Figure~\ref{fig:stats_wind} shows the distribution of wind speed in clear and plume images. We see that the histogram of images with plumes is slightly shifted to the left compared to the histogram of all images which indicates that plumes are better spotted on images with low wind speeds. \\

MARS-S2L is trained for 170 epochs with Adam optimisation \cite{kingma2014adam}, a learning rate of 5e-4, weight regularisation of 1e-6 and early stopping. On each epoch we perform 682 model steps with a batch size of 96. Model selection is performed using mean average precision on the validation split. Validation is always performed on the real data with no simulation. For offshore locations we additionally fine-tune the model for an extra epoch using only real data. We train the model on a single Azure virtual machine with an A100 80Gb Nvidia GPU, 24 CPU cores and 220GB of RAM. Each training run takes approximately 10 hours.  \\

\subsubsection{Comparison to CH4Net and Multi-band multi-pass retrieval}

We emphasize that CH4Net is trained with the full MARS-S2L dataset as opposed to only a small set of emitters in the original paper, and is restricted to the bands available for both Landsat and Sentinel-2.

For comparison to CH4Net, we retrain the model originally developed in \cite{vaughan2023ch4net} using the same training dataset employed for MARS-S2L. Since CH4Net was initially designed for Sentinel-2 imagery and made use of all 13 spectral bands, we adapt the model to use only the six bands that overlap with Landsat. This ensures compatibility with both sensors and enables a fair comparison with MARS-S2L across the full dataset. The excluded bands lie outside the methane absorption region and are therefore not expected to materially impact detection performance. CH4Net is trained for 200 epochs using the simulation scheme described above with Adam optimisation \cite{kingma2014adam}, a learning rate of 5e-4, weight regularisation of 1e-6 and early stopping. We additionally trained CH4Net with the simulation scheme described in previous section~\ref{sec:model_training} as an ablation experiment (see table~\ref{tab:metrics_by_type_of_loc}). We emphasize that this training dataset is considerably larger than that of the original CH4Net paper, therefore this instantiation of CH4Net is significantly better than the original version optimised for Turkmenistan. We additionally note that both CH4Net and the MARS-S2L model use the same UNet architecture with the only difference being the number of input channels.\\ 

For comparison with MBMP retrieval we directly apply a threshold to this product and apply the same procedure using 100 pixels connected components to obtain the scene-level prediction (see section~\ref{sec:class_score}). The MBMP retrieval is an estimation of the atmospheric transmittance hence lower MBMP values indicates higher presence of methane~\cite{gorrono2023understanding}. In particular, for comparisons in tables 6-8 we use 0.99 threshold which provides a similar recall to the MARS-S2L model (see PR and ROC curves~\ref{fig:pr_roc_curves} for full values).

\subsubsection{CloudSEN12+ Experiment}

CloudSEN12+ (Aybar et al.~\cite{cloudsen12,aybar_cloudsen12_2024}) is the largest publicly available dataset for Sentinel-2 cloud detection, featuring 10,440 globally distributed, multi-temporal samples. Each sample includes five Sentinel-2 L1C images with human-annotated cloud labels systematically covering varying cloud-coverage levels (0\% to $>$65\%). We use CloudSEN12+ cloud-free images to assess the false positive rate of the models on areas not covered by the MARS-S2L dataset. We assume that in all CloudSEN12+ images there is no methane emission occurring. This experiment essentially tests the out-of-distribution (OOD) false positive rate (FPR) of the model because CloudSEN12+ contains diverse, globally representative scenes that differ from the MARS-S2L training distribution, testing the model’s generalization potential for global deployment.\\

For each of the 10,440 images we downloaded the corresponding 200x200 Sentinel-2 L1C image and the most similar cloud-free image to be used as a reference within a time window of 8 months. We found a cloud free reference for all images except for 6. We compute the cloud masks of these images using the CloudSEN12 cloud detection model and fetched the wind data from ERA5Land or NASA/GEOS/FP depending on the location. Figure~\ref{fig:cloudsen12_locs} shows the spatial distribution of these images. Table~\ref{tab:fprcloudsen12} shows the false positive rate of the MARS-S2L model together with the two baseline models over this dataset. We see that the MARS-S2L FPR is slightly higher than in the MARS-S2L test set (7.8\%). \\

\subsubsection{Mitigation cases}\label{subsec:mitigation_cases_methods}

This subsection provides details on the four additional successful mitigation cases referenced in Section~\ref{sec:successful_mitigation_cases} (Kazakhstan, Yemen, Argentina, and Turkmenistan), complementing the in-depth discussion of the Algeria and Libya cases. For each case, we document the timeline of satellite-based detections, notification procedures, and the mitigation actions confirmed through follow-up engagement and subsequent monitoring ~\cite{unep_imeo_2024,unep_imeo_2025}.\\

\paragraph{Yemen}
The persistent methane source at the Kamil Field in Yemen was first detected by NASA’s EMIT instrument in August 2022. A subsequent analysis of satellite archives identified 130 plumes from this site in Sentinel-2 and Landsat imagery, with the earliest detection dating to January 2022. Following the appointment of a national focal point in December 2024, a MARS notification was issued summarising the case. The operator’s investigation traced the emissions to a deep casing failure. Temporary production halts and repair attempts proved ineffective, as emissions recurred and were repeatedly confirmed by satellite observations. The well was therefore permanently shut down on 13 April 2025, and mitigation was verified using multiple independent satellite datasets. It is estimated that approximately 19,000 tonnes of methane were released from this source in the year preceding mitigation. Figure \ref{fig:mitigation_yemen} (Top) shows the temporal evolution of the emissions from January 2025, with a temporary cessation followed by a resurgence in mid-April before the final shutdown, while examples of detected plumes are shown in Figure \ref{fig:mitigation_yemen} (Bottom).\\

\paragraph{Kazakhstan}
A persistent methane source in Kazakhstan’s Middle Caspian Basin was first identified through a hyperspectral detection by NASA’s EMIT instrument in June 2023. One year later, the MARS-S2L system detected nine plumes between 19 July and 22 August 2024, with an average emission rate of 2.3 tonnes of methane per hour. Notifications were subsequently sent to the relevant national and company focal points. The operator’s on-site inspection traced the emissions to a malfunctioning valve, which was replaced on 7 April 2025. Satellite observations from multiple instruments confirmed the cessation of emissions following the repair. The timeline of detections from Sentinel-2 and Landsat is shown in Figure~\ref{fig:mitigation_kazakhstan}.\\

\paragraph{Argentina}
At the Meseta Espinosa site in Argentina, the MARS system detected and reported three distinct methane plumes. The first was observed on 25 March 2025, following an earlier, weaker plume on 21 March that was not captured by the model. A final detection on 7 June 2025 was promptly notified, and the operator subsequently confirmed that the emissions originated from an operational failure in a separator, where low temperatures had diverted gas to a storage tank. The issue was resolved on 9 June through repairs to the pneumatic liquid discharge equipment, with subsequent satellite monitoring confirming successful mitigation. The timeline of emissions, including the earlier unmodelled event, along with examples of detected plumes, are shown in Figure~\ref{fig:mitigation_argentina}.

\paragraph{Turkmenistan}
In Turkmenistan, the MARS-S2L system identified recurring methane emissions at the Northern Balguyy site within the Abadanchylyk field, with seven distinct events detected since mid-July 2024. Notifications were issued to the operator in October 2024 and again in May and June 2025. Following the final notification on 3 June 2025, the operator conducted an investigation and attributed the leaks to extensive wear and tear on pipelines connecting several wells to a gas collection point. Maintenance and repair work were subsequently completed, eliminating the source. The timeline of these emission events is shown in Figure~\ref{fig:mitigation_turkmenistan}.

\subsubsection{Calculation of equivalent emissions}
\label{rwi}
To calculate the equivalent emissions for the mitigated source in Hassi Messaoud, we first estimate the total annual methane emissions of this source by taking the average of the estimated hourly flux over all included images in 2024 and multiplying this by 24 (hours in each day) and 365 (days in a year) giving a total of 27,500 tonnes. We calculate $CO_2$ equivalence based on both the 10 and 20 year global warming potential (GWP). The 20 year GWP (GWP-20) of methane is 81 \cite{IPCC_AR6_WG1_Chapter7} giving a $CO_2$ equivalence of 2,227,500 tonnes. Taking the yearly emissions of a standard US car to be 4.6 tonnes \cite{EPA_GHG_Passenger_Vehicle}, this equates to the emissions of roughly 484,000 cars. Similarly, taking the 100 year GWP to be 27.9 \cite{IPCC_AR6_WG1_Chapter7} gives a $CO_2$ equivalence of 767,250 tonnes. This is equivalent to 166,793 standard US cars. To derive comparisons to different countries we compare the $CO_2$ equivalent emissions derived above to the methane emissions of countries globally taken from \cite{OWID_Methane_Emissions}.

\subsection{Carbon Footprint of Model Development and Deployment}
This section presents an estimate of the greenhouse gas emissions associated with compute usage for the machine‑learning models developed in this study, based on energy consumption during model training and its operational deployment. 

During the development of the AI models presented in this paper, we trained fewer than 150 models on a high-performance Azure VM equipped with an NVIDIA A100 GPU. Each training run lasted approximately 18 hours, resulting in an estimated training-related carbon footprint of roughly 2.7 to 3 tons of CO$_2$e, assuming full GPU utilization. The large number of training runs was necessary to explore different combinations of model architectures, input preprocessing strategies, and hyperparameter settings, as well as to iterate on a robust and bug-free training pipeline. Like many real-world ML projects, a significant portion of these runs were part of the experimentation and debugging process needed to reach a reliable and high-performing model.

Once trained, model inference is performed on a standard non-GPU virtual machine (\texttt{Standard\_E4s\_v3} on Microsoft Azure), which runs daily within the MARS ingestion pipeline. On average, this pipeline takes 3 hours each day, though less than 5\% of that time is used for running the AI model—the rest is dedicated to tasks such as downloading raw images, preprocessing, and storing outputs in a database. As a result, the inference workload contributes only about ~2.5 kg of CO$_2$e per month, making its ongoing footprint relatively low compared to training. Looking ahead, we plan to retrain the model every 3 to 4 months to incorporate new data and maintain performance, with each retraining cycle adding approximately 18–20 kg of CO$_2$e. Overall, while training accounts for the vast majority of emissions, the operational phase remains very lightweight and efficient in terms of environmental impact.

\subsection*{Supplementary Text}

\subsubsection*{Extended data}

\normalsize

\begin{figure}
   \centering
\includegraphics[width=1.0\linewidth]{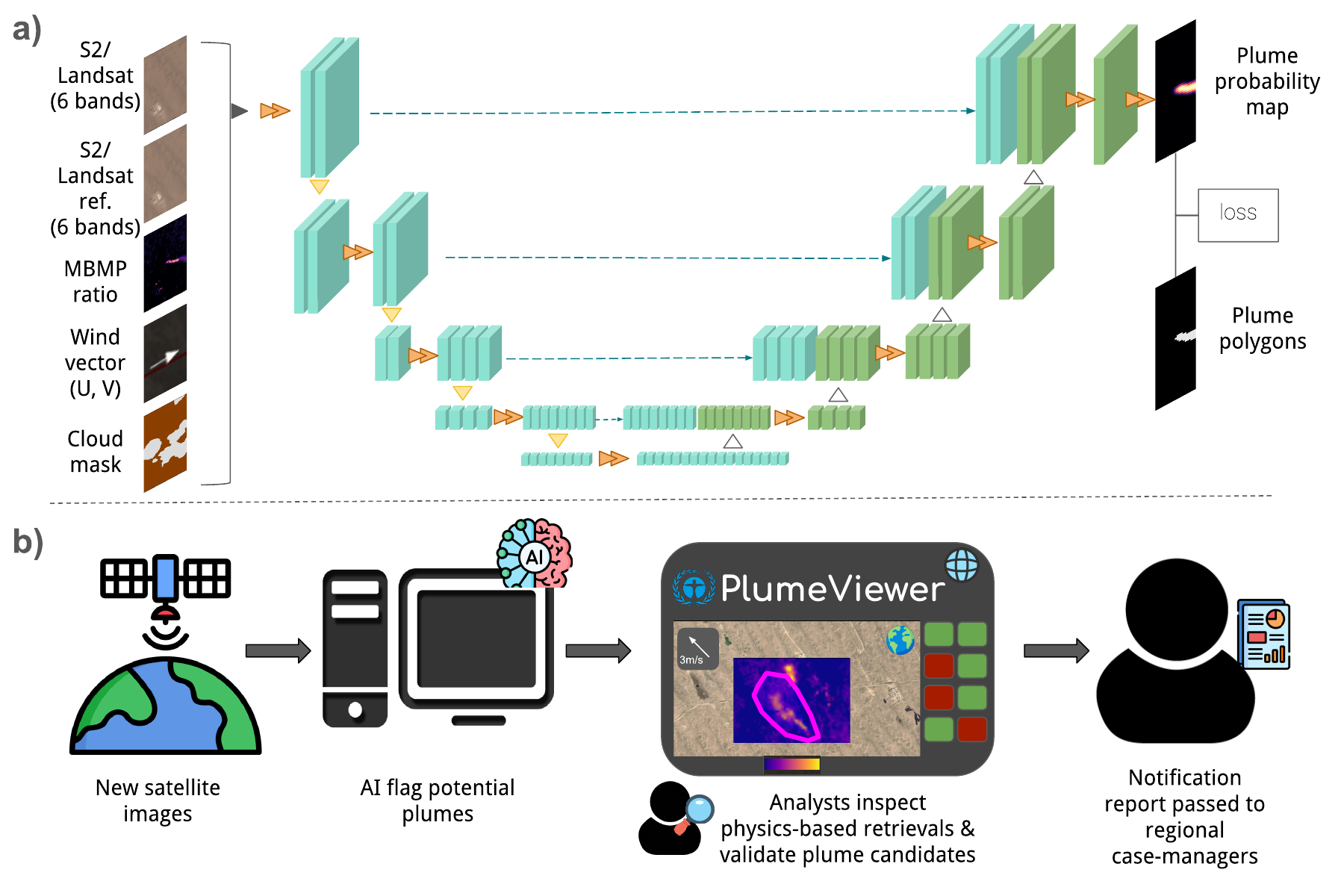}
    \caption{\textit{\textbf{MARS-S2L model architecture and deployment} (a) the architecture and inputs of the MARS-S2L model. The multispectral bands from the current and reference overpass the cloud mask, the MBMP image and wind information are used as inputs. A cloud mask is first generated using the CloudSEN12 model. All data is then fed into MARS-S2L which outputs the probability that each pixel is part of a methane plume. (b) shows the operational deployment process. At 06:30 every morning any new Sentinel-2 and Landsat images are downloaded and predictions generated. These are then shown in the PlumeViewer where analysts inspect each alert and provide details of events over known assets to case managers to issue notifications.}}
    \label{fig:modelandplumeviewer}
\end{figure}

\begin{figure}
    \centering    \includegraphics[width=0.63\linewidth]{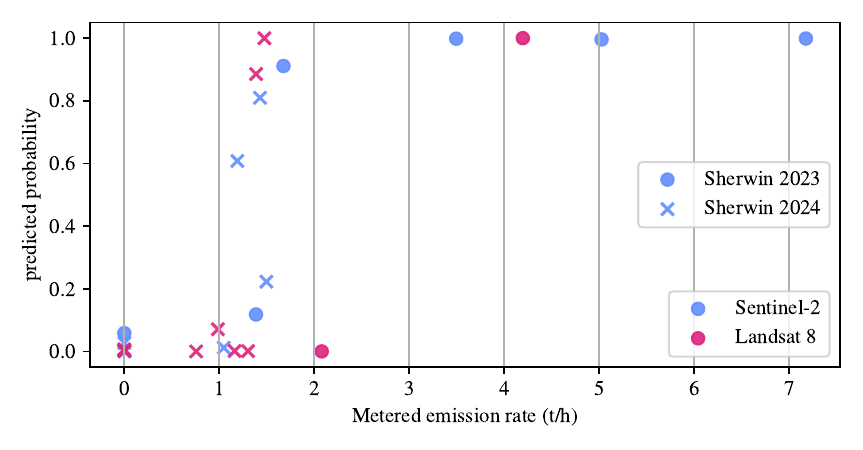}\includegraphics[width=0.33\linewidth]{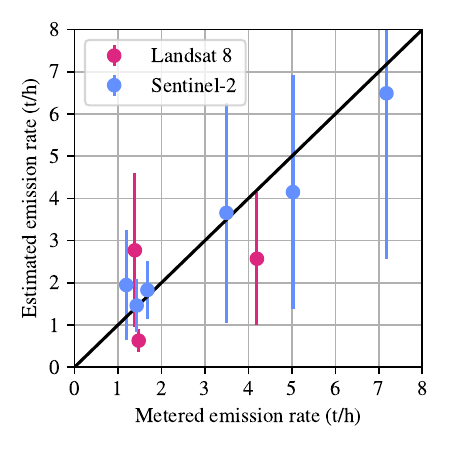}
    \caption{\textit{\textbf{Evaluation of the model over controlled releases}. \textbf{Left} predicted probability of the model in the controlled release experiments of Sherwin et al.~\cite{sherwin_single-blind_2023,sherwin_single-blind_2024}. We see that the model assigns near-zero probability to all no emission samples in the experiment. \textbf{Right}: quantified emissions of positive detected samples using the mask provided by the MARS-S2L model. }}
    \label{fig:controlled_releases}
\end{figure}

\begin{figure}
    \centering
    \includegraphics[width=0.8\linewidth]{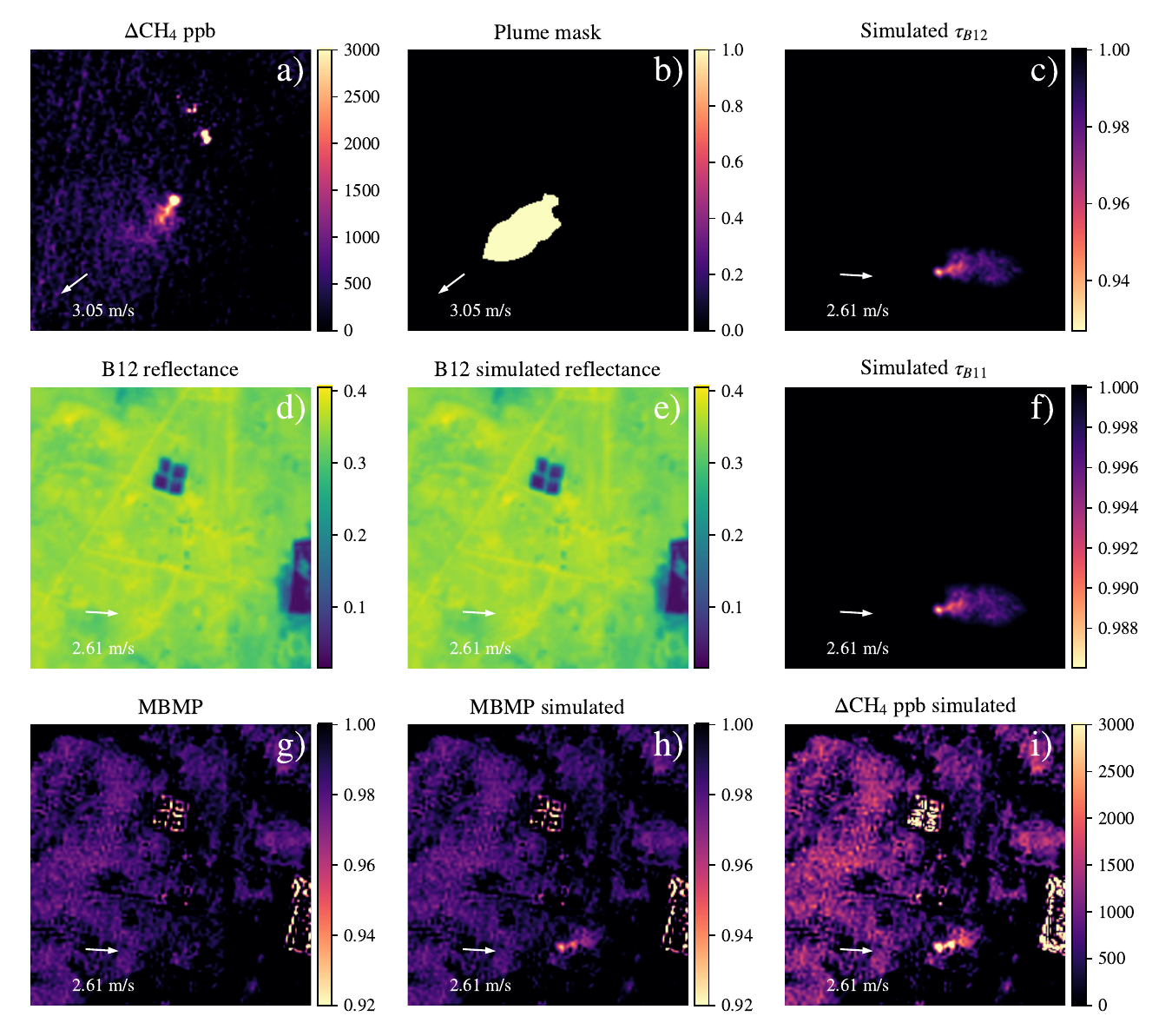}
    \caption{Samples highlighting the simulation procedure: (a) quantified retrieval $\Delta$CH$_4$ of a plume, (b) binary plume mask for the same plume, (d) B12 reflectance and (g) MBMP for an image in a different location without a plume. Transmittance of the plume in bands 11 (f) and 12 (c) aligned with the wind of the clear location, B12 band of the clear image with the simulated plume (e), MBMP retrieval (h) and quantified retrieval (i).}
    \label{fig:simulation_images}
\end{figure}

\begin{figure}
    \centering
    \includegraphics[width=0.7\linewidth]{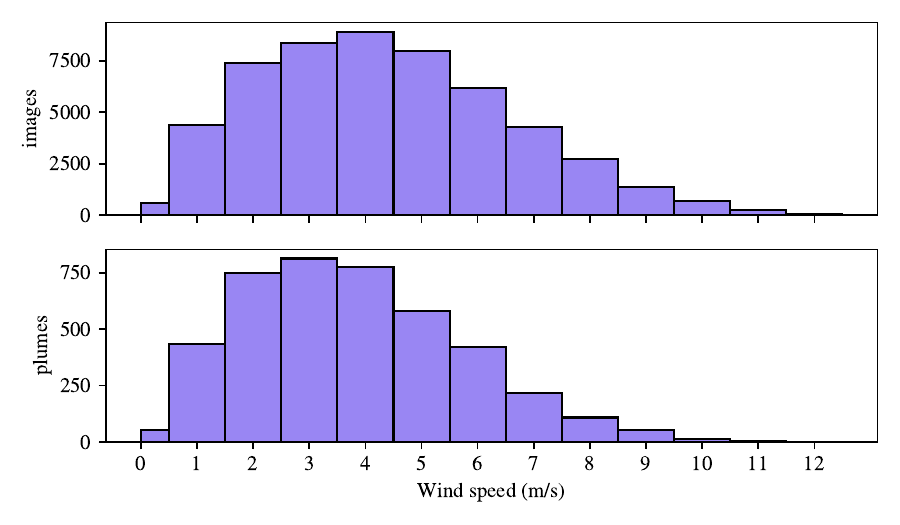}
    \caption{Wind speed distribution in the MARS-S2L dataset. \textbf{Top} wind speed in all images. \textbf{Bottom} wind speed in images with validated emissions. Plumes are simulated in clear images based on their wind speed: the plume to be simulated must have a wind speed difference of less than 1.5m/s with the wind of the clear image. In addition we do not simulate plumes in images with more than 9m/s wind speed.}
    \label{fig:stats_wind}
\end{figure}

\begin{figure}
    \centering
    \includegraphics[width=0.8\linewidth]{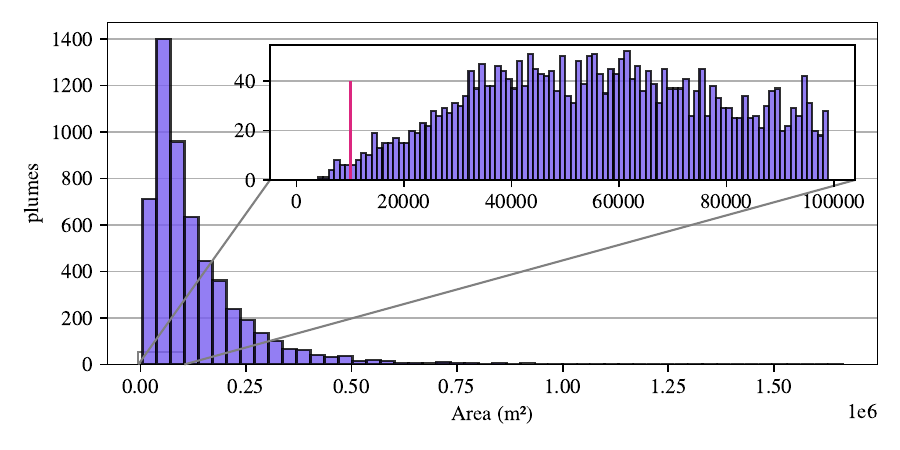}
    \caption{Distribution of the area labeled of the plume in square meters. The red vertical line shows the 100 pixel threshold (0.01 km²) used to produce the scene level metrics (see section~\ref{sec:methods}).}
    \label{fig:areaplume}
\end{figure}

\begin{figure}
    \centering
    \includegraphics[width=0.95\linewidth]{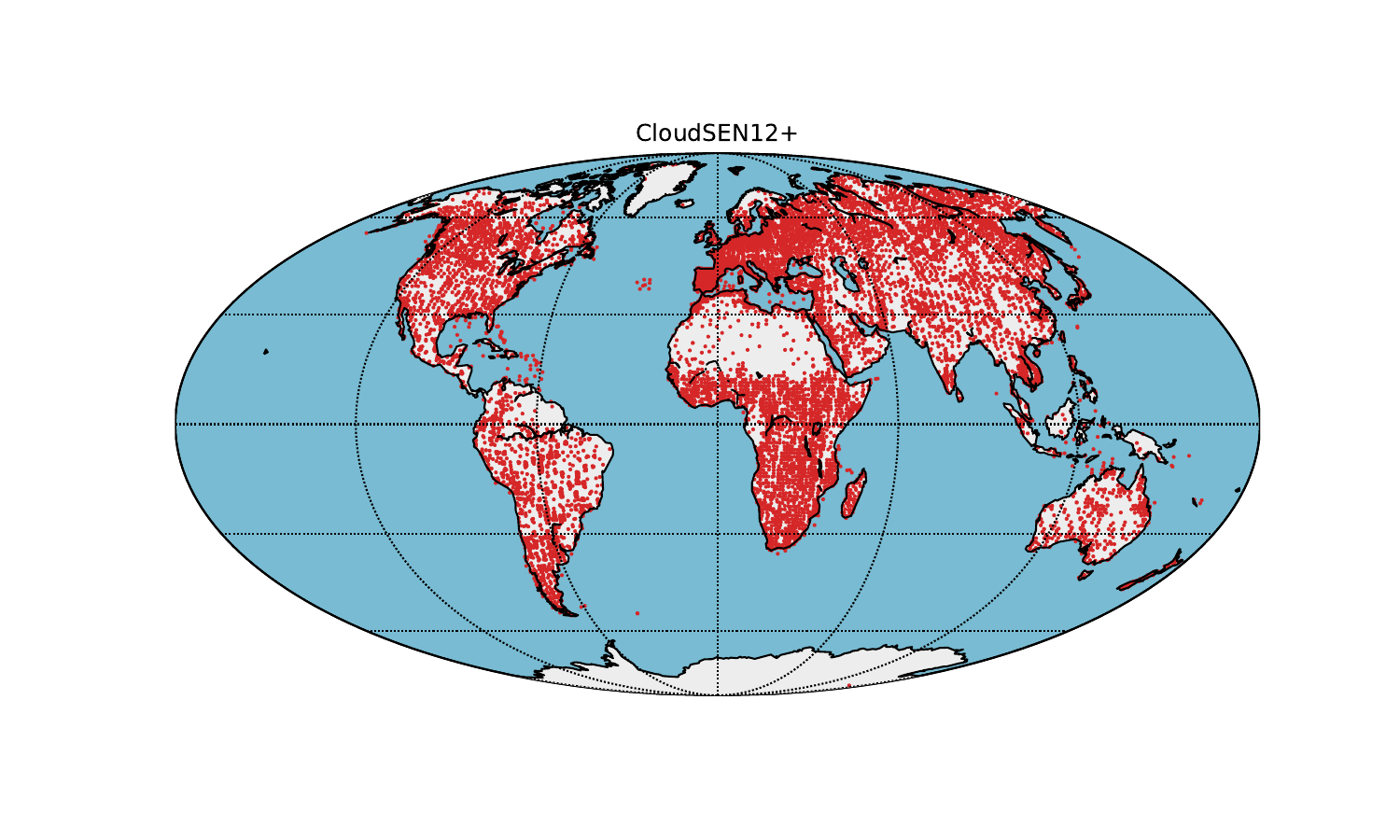}
    \caption{Distribution of CloudSEN12+ images used to estimate the global false positive rate across different background surfaces and atmospheric conditions.}
    \label{fig:cloudsen12_locs}
\end{figure}

\begin{figure}
    \centering
    \includegraphics[width=\linewidth]{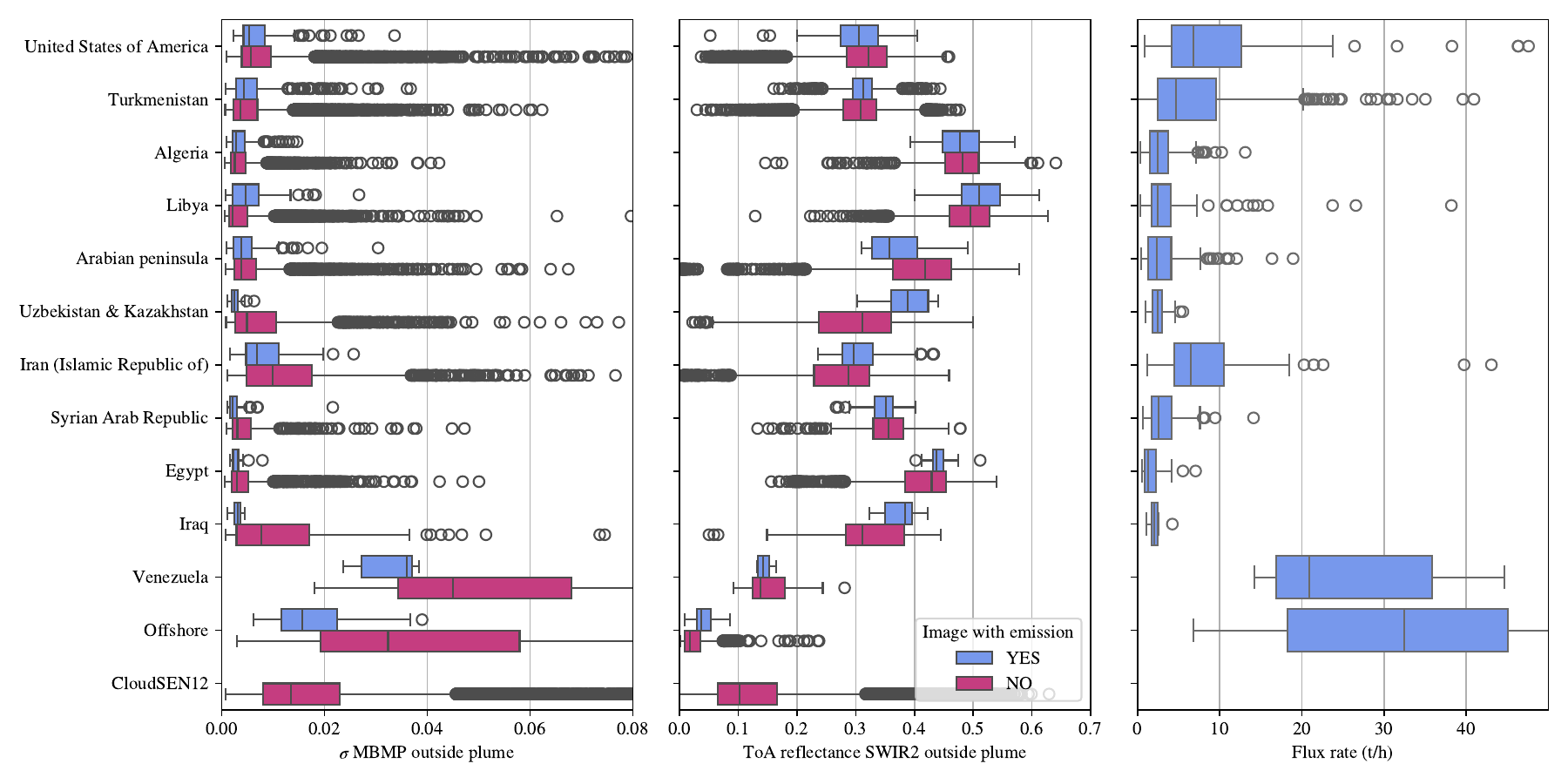}\\
    \caption{Distribution of image statistics across the selected case studies on the MARS-S2L test dataset. \textbf{Left} Distribution of the standard deviation of MBMP. \textbf{Center} distribution of average ToA reflectance in SWIR2 band. \textbf{Right} Distribution of flux rates of plumes detected. Plumes are better detected in images with high reflectance and low MBMP difference.}
\label{fig:stats_marss2l_test}
\end{figure}

\begin{figure}
    \centering
\includegraphics[width=0.53\linewidth]{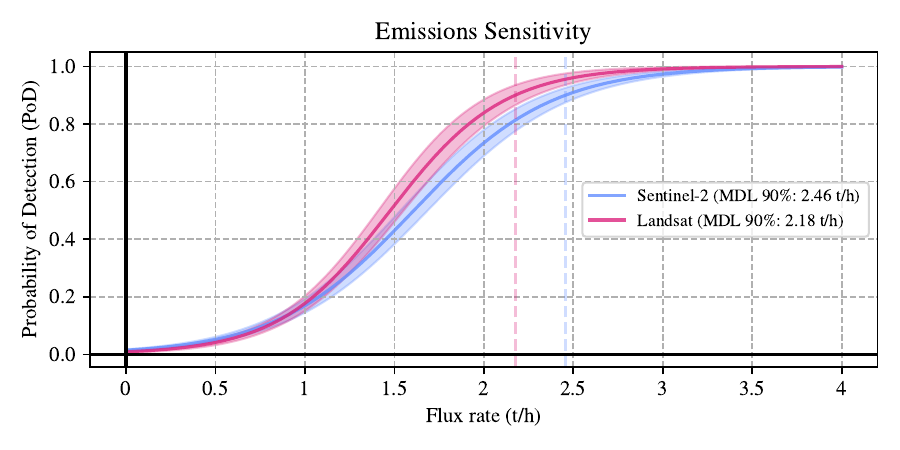}
    \includegraphics[width=0.35\linewidth]{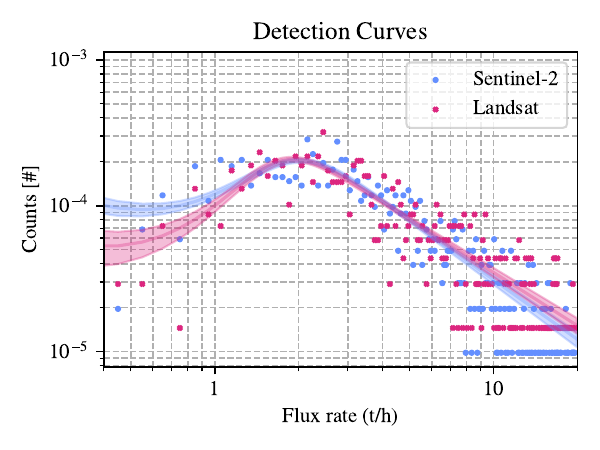}\\
    \includegraphics[width=0.53\linewidth]{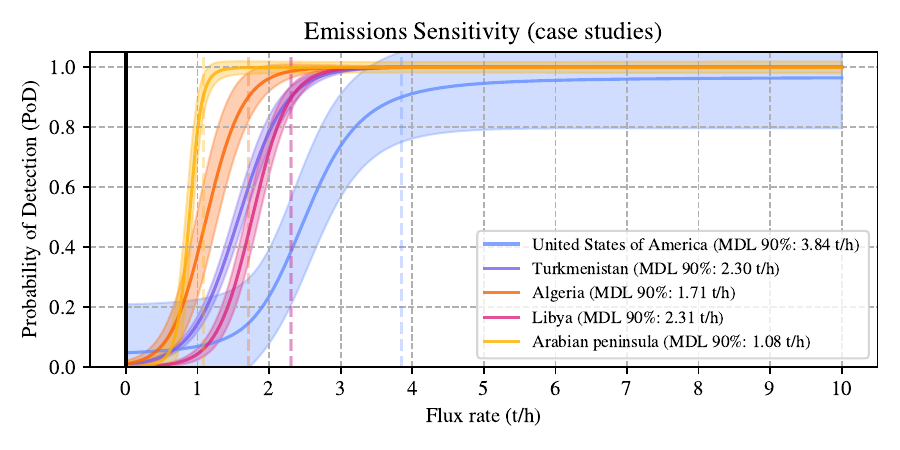}
    \includegraphics[width=0.35\linewidth]{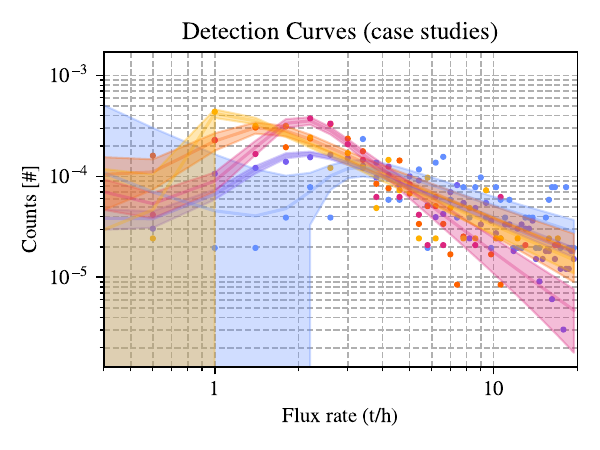}\\

    \caption{Observational probability of detection curves estimated in the MARS-S2L test dataset following the methodology of Ethret et al.~\cite{ehret_global_2022}. \textbf{Top}: Overall sensitivity for Sentinel-2 and Landsat in all arid and semi-arid regions (all case study areas except offshore and Venezuela). \textbf{Bottom}: Sensitivity in the case studies with more than 100 plumes in the test set. These curves estimate the probability of detection of plumes of UNEP analysts based on the existing observational data of 2024.}
    \label{fig:mdl_PoD}
\end{figure}

\begin{figure}
    \centering
    \includegraphics[width=.9\linewidth]{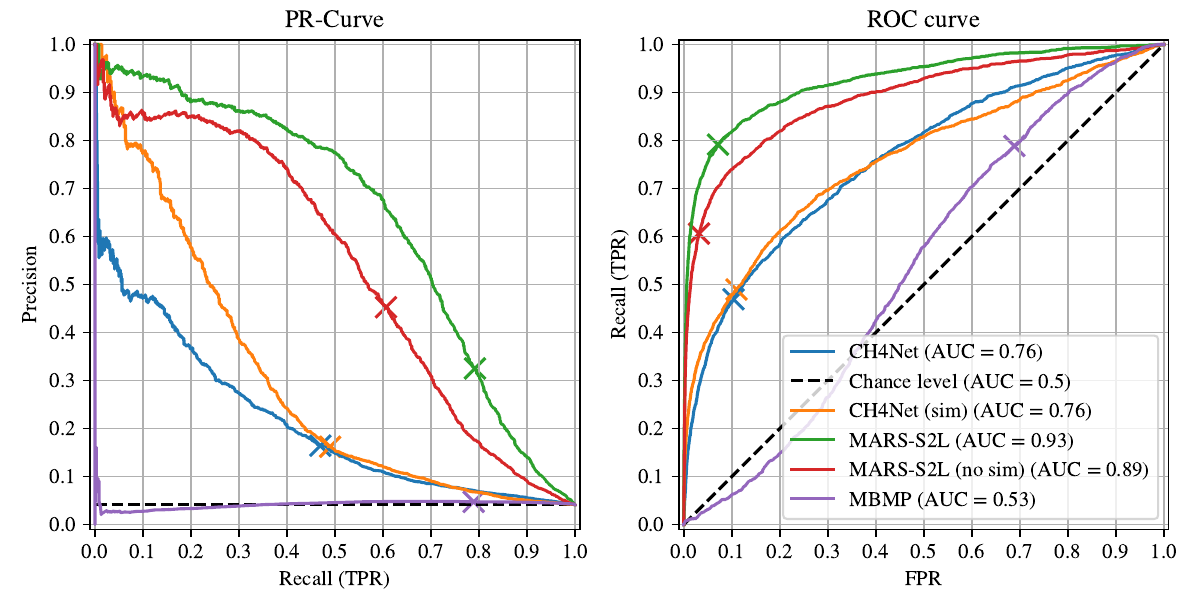}
    \caption{Precision-Recall and receiver operator (ROC) curves of the models on MARS-S2L test dataset. The crosses show the metrics of the selected threshold of the models (0.5 for CH4Net and MARS-S2L and 0.99 for MBMP). These crosses correspond to the statistics shown in Table~\ref{tab:overall_metrics}.}
    \label{fig:pr_roc_curves}
\end{figure}

Here we present extended data images for the methodology section, showing the model architecture and operational pipeline \ref{fig:modelandplumeviewer}, simulation process (Figure \ref{fig:simulation_images}), wind speed distribution in the MARS-S2L dataset to inform the simulation procedure (Figure \ref{fig:stats_wind}) 
and results on the controlled release experiments (Figure \ref{fig:controlled_releases}). We also include the distribution of images in the CloudSEN12+ experiment (Figure~\ref{fig:cloudsen12_locs}, the overall statistics obtained in this experiment~\ref{tab:fprcloudsen12} and the table with top 25 countries with the highest percentage of upstream methane emissions by according to IEA~\cite{iea2025methane} (Table~\ref{tab:percentage_emissions_iea}). 

\newpage

\begin{table}[ht]
\centering
\footnotesize
\begin{tabular}{lrr}
\toprule
Model & Threshold & FPR (\%) \\
\midrule
MARS-S2L & 0.50 & 9.43 \\
CH4Net & 0.50 & 1.97 \\
MBMP & 0.99 & 92.98 \\
\hline
MARS-S2L & 0.90 & 1.81 \\
CH4Net & 0.90 & 0.58 \\
MBMP & 0.90 & 27.65 \\
\hline
MARS-S2L & 0.98 & 0.52 \\
CH4Net & 0.98 & 0.24 \\
MBMP & 0.85 & 16.64 \\
\bottomrule
\end{tabular}
\caption{False positive rate of plume detection models over 10,434 clear images of CloudSEN12+ for different score thresholds.}
\label{tab:fprcloudsen12}
\end{table}

\begin{table}[]
    \centering
    \footnotesize
    \begin{tabular}{llr}
\toprule
Country & MARS-S2L Case study area & Methane emissions (\%) \\
\midrule
Offshore & Offshore & 22.88 \\
United States & United States of America & 15.44 \\
Russia & - & 14.52 \\
Iran & Iran (Islamic Republic of) & 5.56 \\
Algeria & Algeria & 3.92 \\
Venezuela & Venezuela & 3.87 \\
Iraq & Iraq & 3.44 \\
Canada & - & 3.36 \\
China & - & 3.00 \\
Turkmenistan & Turkmenistan & 2.92 \\
Libya & Libya & 2.39 \\
Saudi Arabia & Arabian peninsula & 2.04 \\
Kazakhstan & Uzbekistan \& Kazakhstan & 1.80 \\
Kuwait & Arabian peninsula & 1.46 \\
Argentina & - & 1.44 \\
United Arab Emirates & Arabian peninsula & 1.26 \\
Nigeria & - & 1.04 \\
Uzbekistan & Uzbekistan \& Kazakhstan & 0.97 \\
Oman & Arabian peninsula & 0.95 \\
Indonesia & - & 0.85 \\
Mexico & - & 0.60 \\
Syria & Syrian Arab Republic & 0.58 \\
Pakistan & - & 0.56 \\
Egypt & Egypt & 0.51 \\
Colombia & - & 0.50 \\
\bottomrule
\end{tabular}
    \caption{Percentage of upstream methane emissions of top 25 most emitting countries (95.85\% of total emissions) according to IEA estimations~\cite{iea2025methane}. MARS-S2L case-studies cover 70\% of these areas.}
    \label{tab:percentage_emissions_iea}
\end{table}

\begin{table}[]
    \centering
    \begin{tabular}{llrrrll}
\toprule
 &  & total  & total & total & min date & max date \\
split & Satellite & images & plumes & sites &  &  \\
\midrule
\multirow[t]{2}{*}{train} & Landsat & 6,685 & 913 & 537 & 2018-01-05 10:02 & 2023-11-30 07:17 \\
 & Sentinel-2 & 31,681 & 2,520 & 598 & 2018-01-01 09:13 & 2023-11-30 10:33 \\
\cline{1-7}
\multirow[t]{2}{*}{val} & Landsat & 262 & 32 & 36 & 2021-01-04 10:08 & 2021-12-31 10:03 \\
 & Sentinel-2 & 5,772 & 256 & 89 & 2021-01-01 07:03 & 2021-12-31 10:03 \\
\cline{1-7}
\multirow[t]{2}{*}{test} & Landsat & 18,247 & 743 & 1,251 & 2024-01-01 00:05 & 2024-12-31 17:27 \\
 & Sentinel-2 & 25,282 & 1,070 & 1,247 & 2024-01-01 03:41 & 2024-12-31 17:06 \\
\bottomrule
\end{tabular}
    \caption{Training, test and validation dataset statistics. Note: year 2021 is excluded from the Train split.}
    \label{tab:splits}
\end{table}

\begin{table}[]
    \centering
    \begin{tabular}{lrrr}
\toprule
 & images & plumes & sites \\
\midrule
Sites seen at training time & 27,874 & 1,586 & 592 \\
Test only sites & 15,655 & 227 & 697 \\
\bottomrule
\end{tabular}
    \caption{Test dataset statistics split by new sites and sites seen at training time. Roughly 35\% of images are from locations not seen at training time. See table~\ref{tab:metrics_by_type_of_loc} for model performance in this split.}
    \label{tab:newsitesoldsites}
\end{table}


\begin{table}[h]
    \centering
    \footnotesize
    \begin{tabular}{llrrrrr}
\toprule
Location & Model & AP   & Precision & Recall & Acc.  & FPR \\
type &  & (\%) & (\%) & (\%) &  (\%) & (\%) \\
\midrule
\multirow{5}{*}{Seeing at training} & MARS-S2L & 68.82 & 41.17 & 79.38 & 92.37 & 6.84 \\
 & MARS-S2L (no sim) & 60.76 & 55.20 & 62.61 & 94.98 & 3.07 \\
 & CH4Net (sim) & 38.86 & 23.71 & 52.90 & 87.63 & 10.27 \\
 & CH4Net & 31.59 & 23.50 & 51.51 & 87.70 & 10.11 \\
 & MBMP & 5.70 & 6.37 & 80.26 & 31.72 & 71.21 \\
 \hline
\multirow{5}{*}{Test only} & MARS-S2L & 44.96 & 13.01 & 77.53 & 92.16 & 7.63 \\
 & MARS-S2L (no sim) & 27.25 & 16.88 & 46.70 & 95.89 & 3.38 \\
 & CH4Net (sim) & 3.28 & 2.53 & 21.59 & 86.82 & 12.22 \\
 & CH4Net & 2.19 & 2.00 & 14.98 & 88.13 & 10.79 \\
 & MBMP & 1.38 & 1.54 & 69.16 & 35.58 & 64.91 \\
\bottomrule
\end{tabular}
\caption{Classification metrics: average precision (AP), recall, precision and FPR on the MARS-S2L dataset stratified by type of location. For comparison purposes, we include the MARS-S2L model trained without simulation and CH4Net with simulation (see section~\ref{sec:model_training}). Table~\ref{tab:newsitesoldsites} shows the statistics on number of images, plumes and sites on each subset (seen at training and test only).}
\label{tab:metrics_by_type_of_loc}
\end{table}

\begin{table}[h]
    \centering
    \footnotesize
    \begin{tabular}{lrrrrr|rrrrr}
\toprule
Model & AP & Precision & Recall & Acc. & FPR & Segmentation & Segmentation & Segmentation & Segmentation & IoU \\
 & (\%) & (\%) & (\%) & (\%) & (\%) & Precision (\%) & Recall (\%) & Acc. (\%) & FPR (\%) & (\%) \\
\midrule
MARS-S2L & 64.08 & 32.53 & 79.15 & 92.29 & 7.13 & 39.93 & 62.60 & 99.78 & 0.16 & 32.24 \\
CH4Net & 21.20 & 16.44 & 46.94 & 87.86 & 10.37 & 15.93 & 23.12 & 99.67 & 0.20 & 10.41 \\
MBMP & 4.22 & 4.74 & 78.87 & 33.11 & 68.88 & 0.26 & 3.21 & 97.84 & 2.00 & 0.24 \\
\bottomrule
\end{tabular}
    \caption{Overall classification and segmentation metrics across the entire MARS-S2L test dataset. We report classification metrics as they are more meaningful for our existing application. Classification scores are derived from segmentation masks as explained in section~\ref{sec:class_score}}
    \label{tab:overall_metrics}
\end{table}

\begin{table}[]
    \centering
    \begin{tabular}{lrrrr}
\toprule
Threshold pixels ($k$) & AP (\%) & Precision (\%) & Recall (\%) & FPR (\%) \\
\midrule
25 & 64.28 & 28.76 & 81.00 & 8.97 \\
50 & 64.58 & 30.56 & 80.50 & 8.18 \\
75 & 64.84 & 31.84 & 79.54 & 7.61 \\
100 & 64.79 & 33.10 & 78.97 & 7.14 \\
125 & 64.79 & 34.29 & 78.18 & 6.70 \\
150 & 64.73 & 35.36 & 77.85 & 6.36 \\
175 & 64.72 & 36.34 & 77.17 & 6.04 \\
\bottomrule
\end{tabular}
    \caption{Metrics of the MARS-S2L model on onshore test locations using different thresholds ($k$) to produce the scene-level scores. The scene-level score is calculated as the minimum threshold of the probability map such that there are at least $k$ connected pixels in the output mask (see Section~\ref{sec:class_score}). We see that changing $k$ the metrics of the model are relatively similar. The selected threshold for the rest of the metrics presented in this paper is 100.}
    \label{tab:scenelevelmetrics}
\end{table}

\subsubsection*{The Plumeviewer}

Analysts logging into the PlumeViewer are presented with recent model detections, which they can inspect to validate plumes, quantify emissions with uncertainty estimates, and cross-reference against infrastructure layers, proprietary databases, and prior detections. Within the PlumeViewer, AI model outputs are manually verified against physics-based Multi-Band Multi-Pass retrieval (MBMP)~\cite{varon2021high,gorrono2023understanding} and several other views of the bands of Sentinel-2 and Landsat to discard false positives. If the detection is recent and can be attributed to a facility in the ground, a formal notification is issued to the government of the country and operator as described in Section 2.3. \\

\begin{figure}
    \centering
    \includegraphics[height=.35\paperheight]{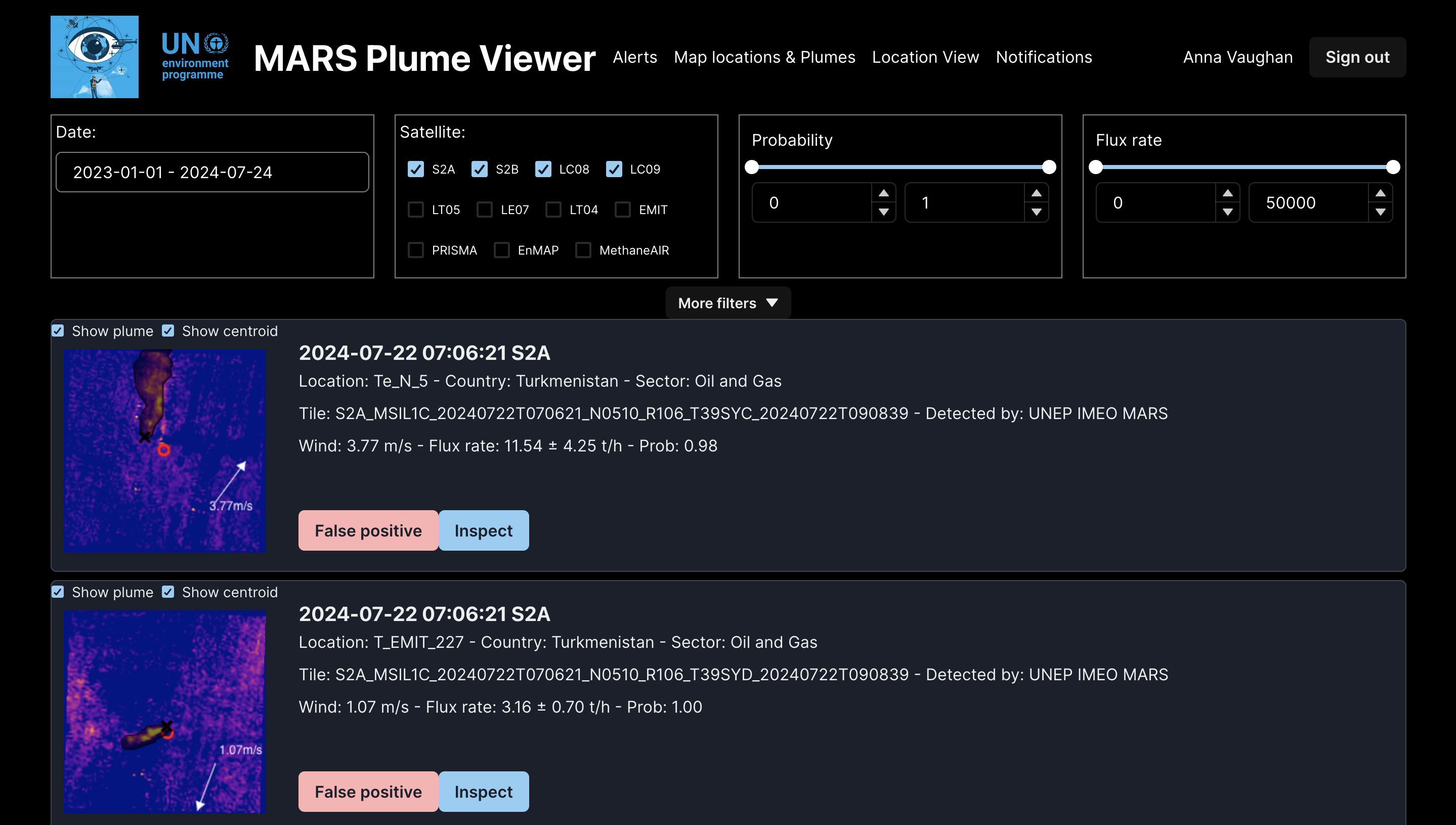}\\
    \caption{PlumeViewer alert view showing model predictions over recently acquired satellite images. }
    \label{fig:plumeviewer1}
\end{figure}

\begin{figure}
    \centering
    \includegraphics[height=.35\paperheight]{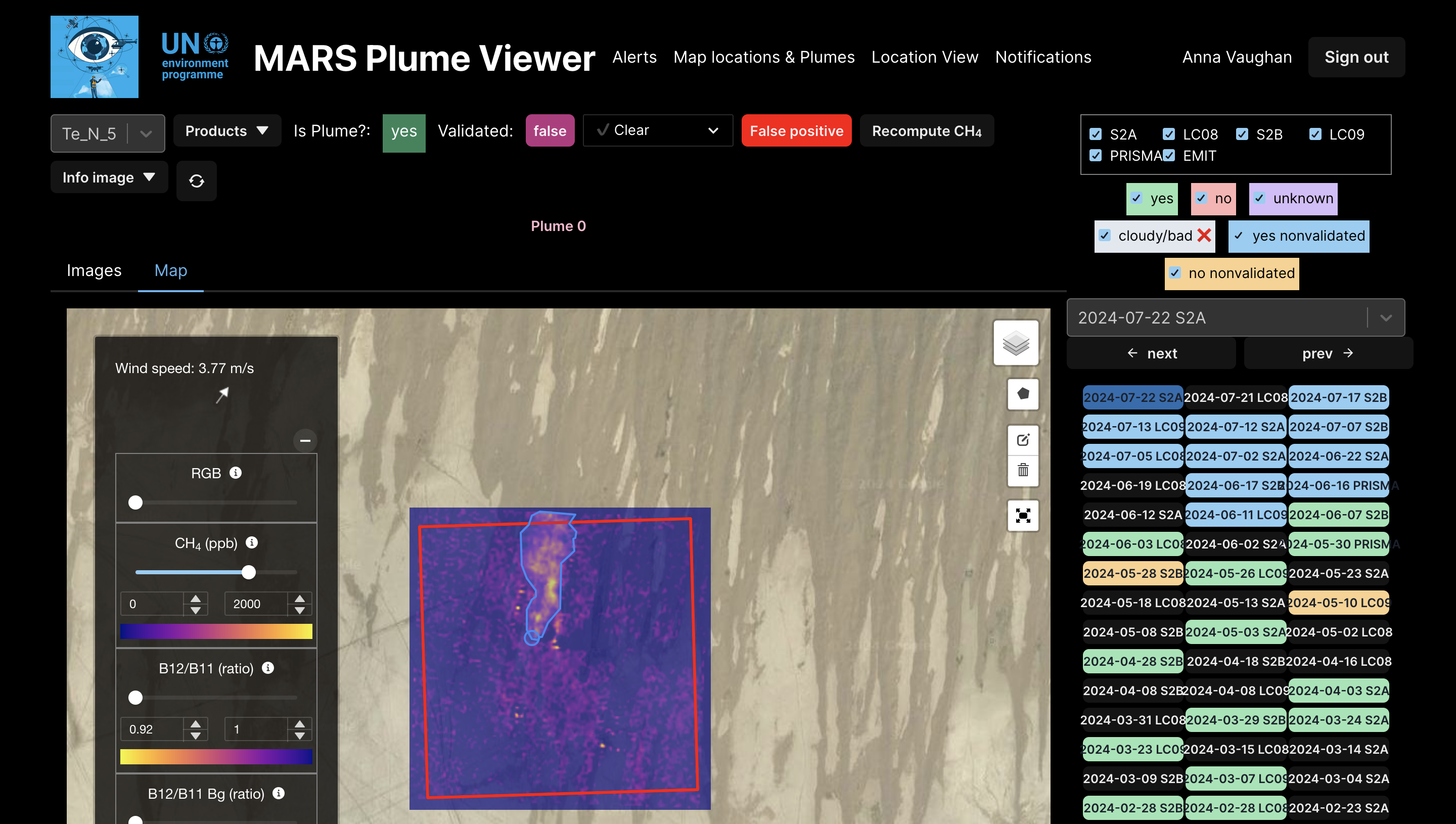}\\
    \caption{PlumeViewer location view showing a model alert for a site.}
    \label{fig:plumeviewer2}
\end{figure}

\begin{figure}
    \centering
    \includegraphics[height=.35\paperheight]{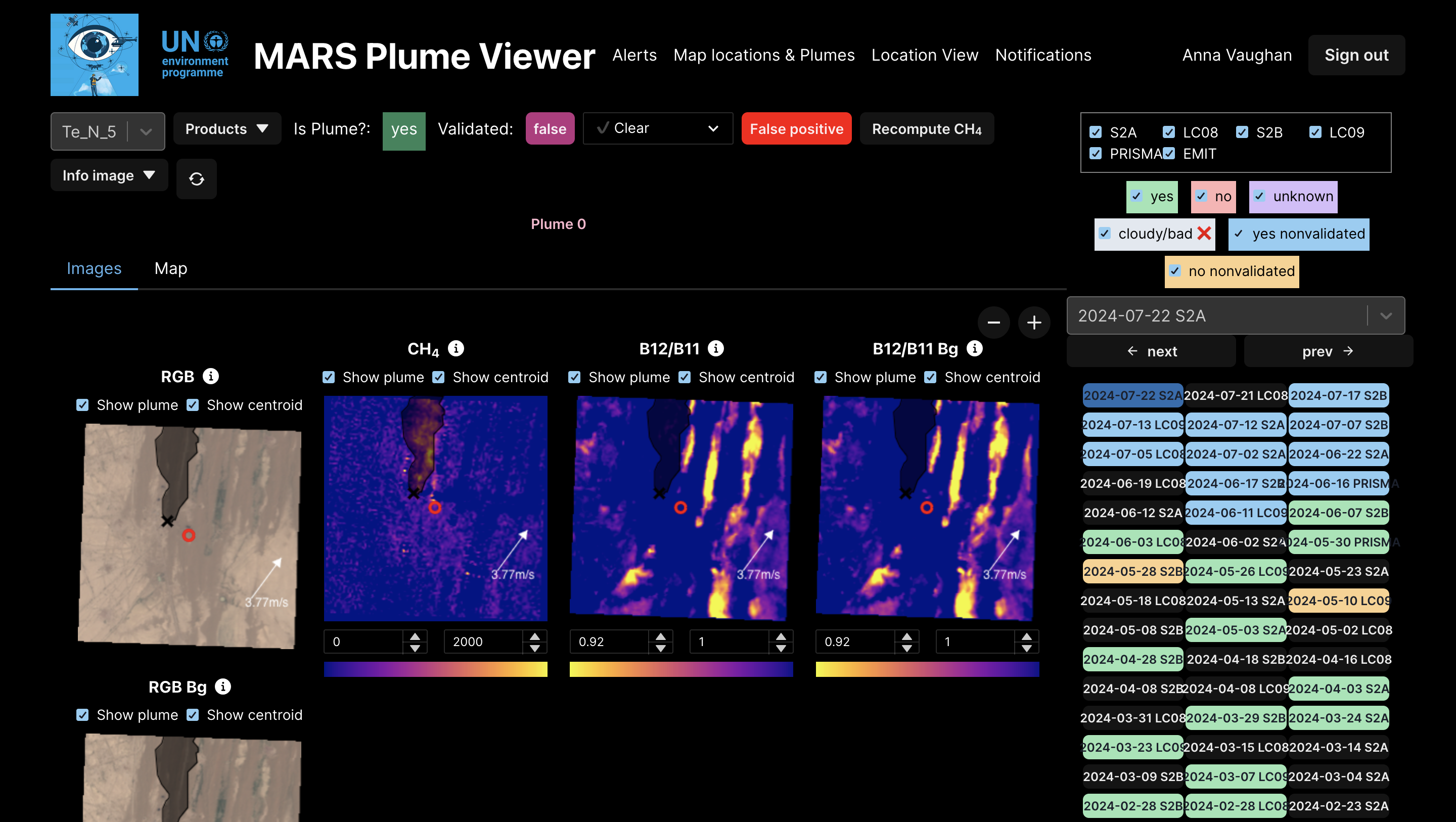}\\
    \caption{PlumeViewer location view showing the quantified plume (figured with CH4 caption) and auxiliary images for a site to aid validation. These auxiliary images include the RGB of the current acquisition, the RGB of the reference acquisition (Bg) and the SWIR2/SWIR1 ratios of the two images.}
    \label{fig:plumeviewer3}
\end{figure}

Figures \ref{fig:plumeviewer1}, \ref{fig:plumeviewer2} and \ref{fig:plumeviewer3} show the three stages of an analyst inspecting an alert produced by MARS-S2L. Figure \ref{fig:plumeviewer1} shows the alert screen where the analyst can inspect model predictions. Any of these alerts can be selected for verification (Figure \ref{fig:plumeviewer2}) with multiple different auxiliary images available for inspection (Figure \ref{fig:plumeviewer3}). \\

\newpage

\subsubsection*{Mitigation cases}
\begin{figure}
    \centering
    \includegraphics[width=0.8\linewidth]{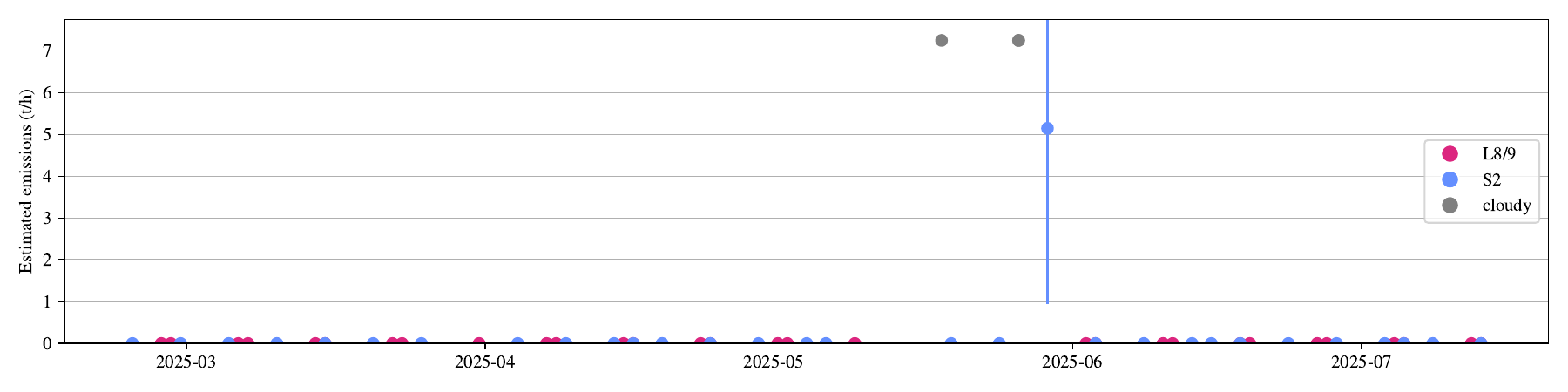}\\
    \includegraphics[width=0.8\linewidth]{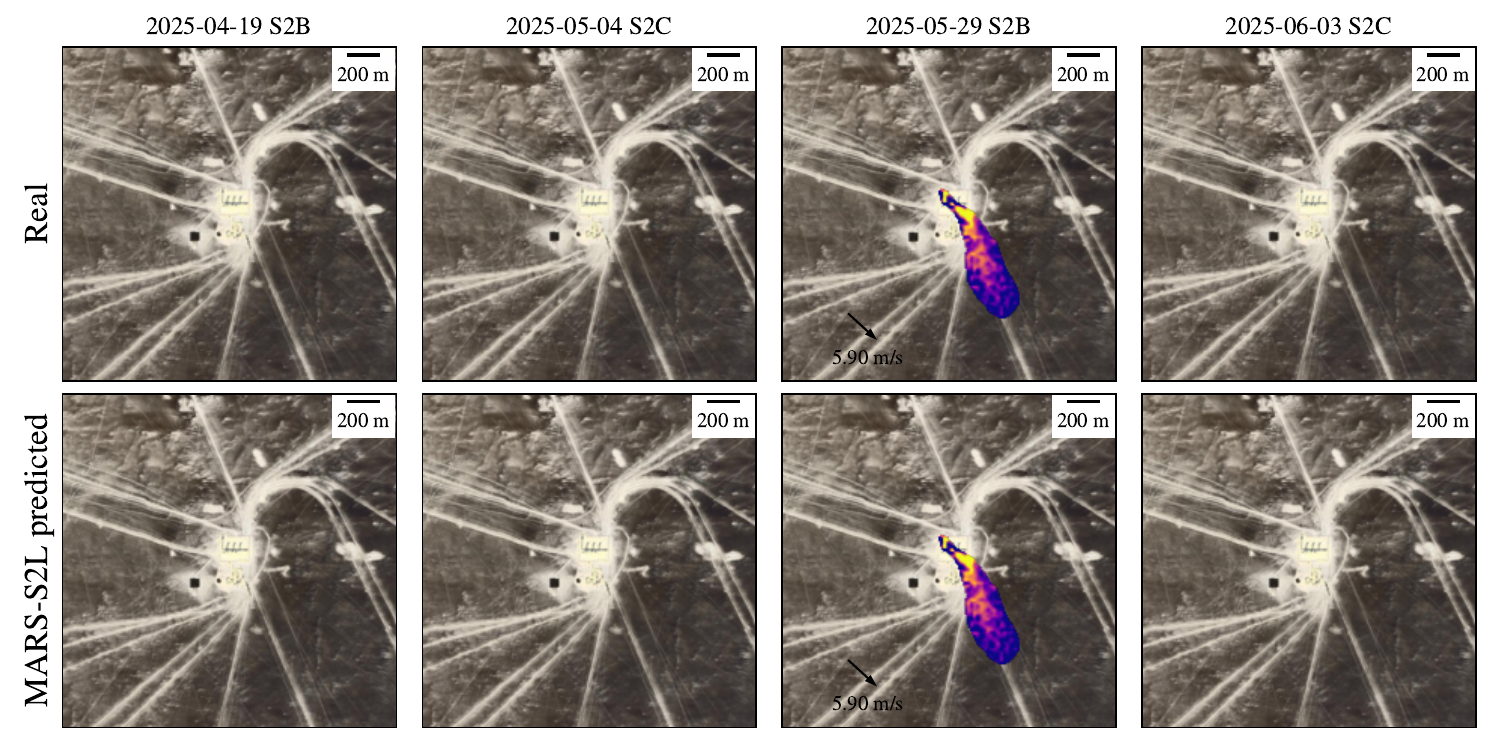}\\
    \includegraphics[width=0.2\linewidth]{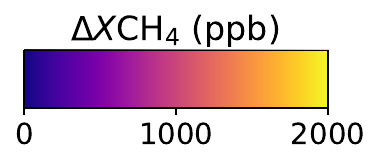}
    \caption{\textbf{MARS-S2L predictions over a mitigated emitter in Libya, Faregh-Argub}. Top: a timeseries of validated emissions with quantification. Bottom: Examples of MARS-S2L predictions compared to expert-annotated ground-truth.}
    \label{fig:mitigation_libya}
\end{figure}

\begin{figure}
    \centering
    \includegraphics[width=0.8\linewidth]{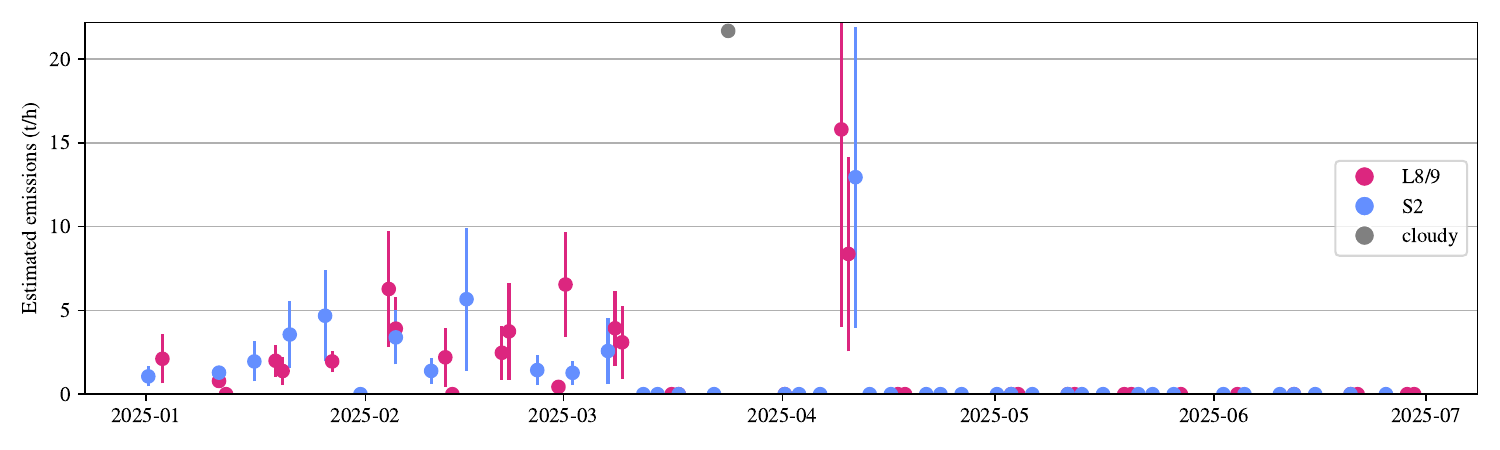}\\
    \includegraphics[width=0.8\linewidth]{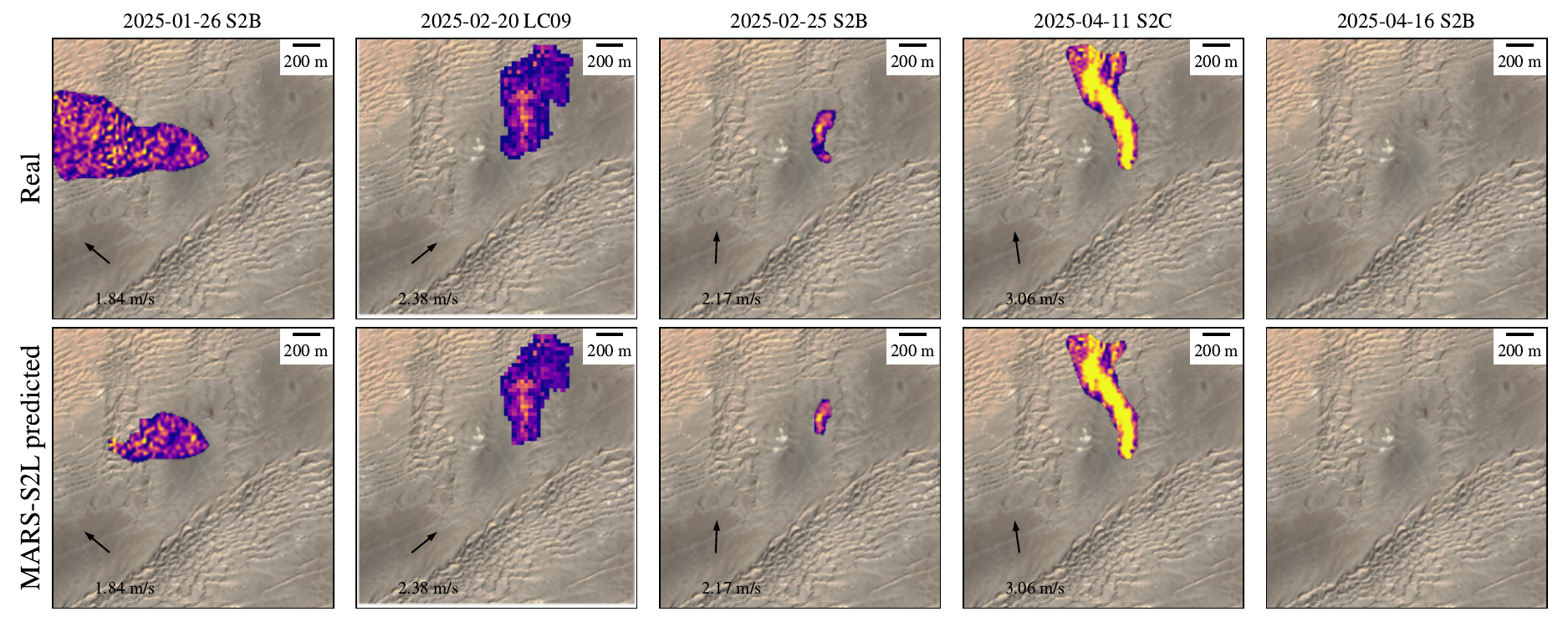}\\
    \includegraphics[width=0.2\linewidth]{figsmitigation/pretty_colorbar_horizontal_2000.pdf}
    \caption{\textbf{MARS-S2L predictions over a mitigated emitter in Yemen, Kamil Field}. Top: a timeseries of validated emissions with quantification. Bottom: Examples of MARS-S2L predictions compared to expert-annotated ground-truth.}
    \label{fig:mitigation_yemen}
\end{figure}

\begin{figure}
    \centering
    \includegraphics[width=0.8\linewidth]{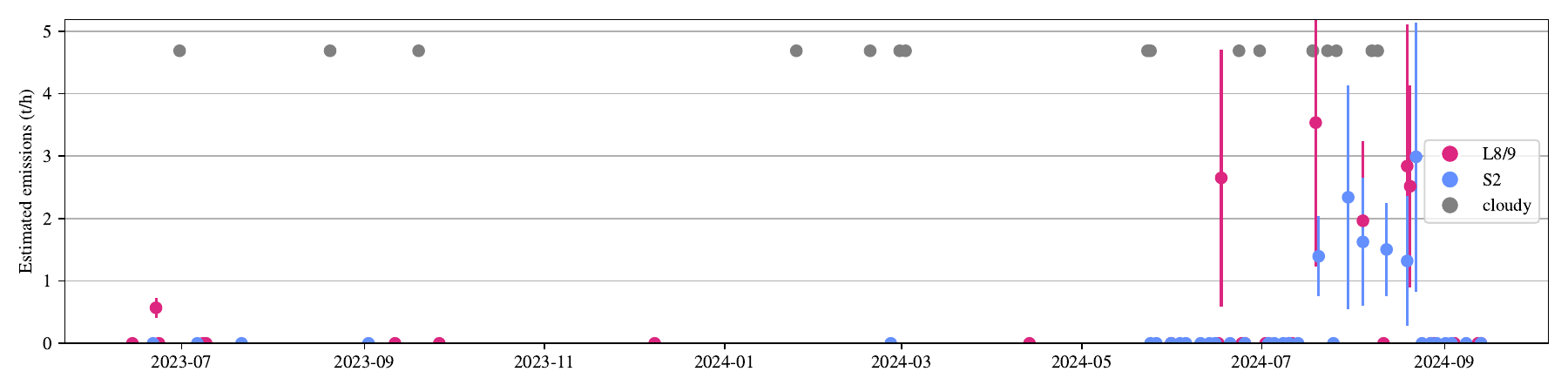}\\
    \includegraphics[width=0.8\linewidth]{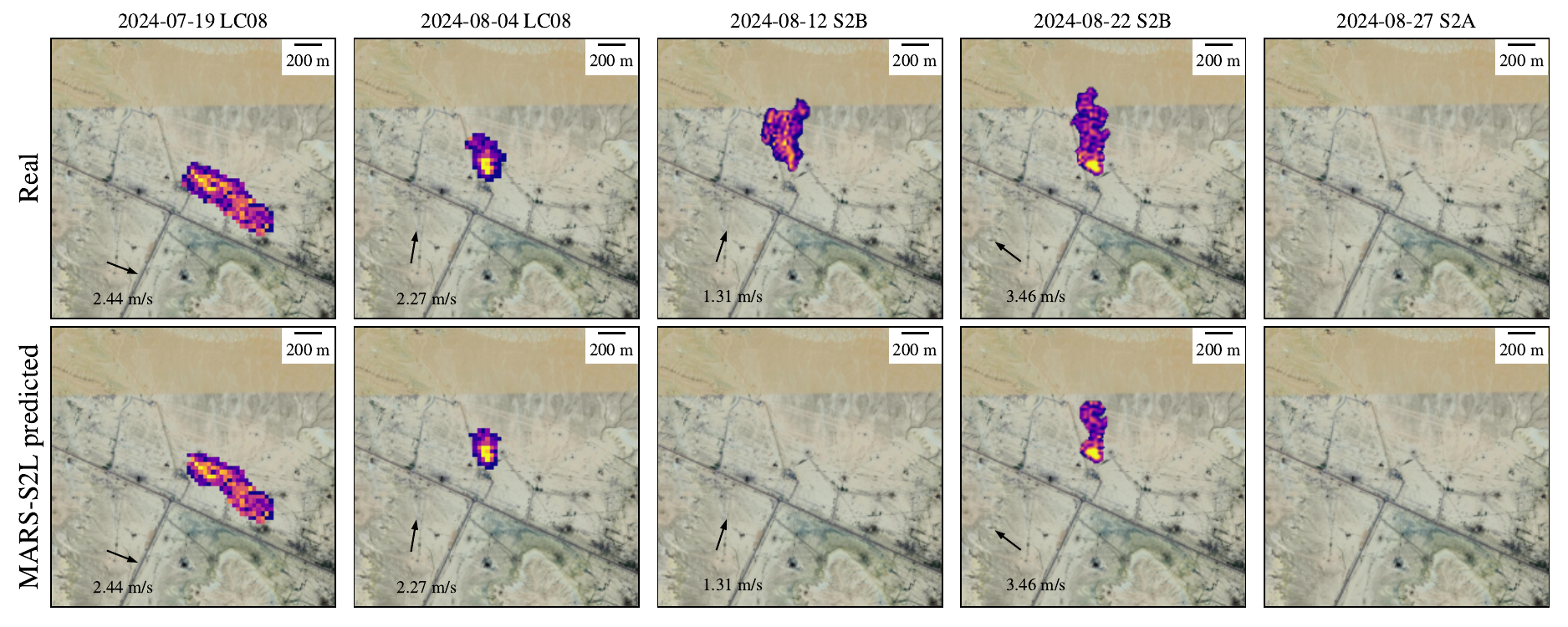}\\
    \includegraphics[width=0.2\linewidth]{figsmitigation/pretty_colorbar_horizontal_2000.pdf}
    \caption{\textbf{MARS-S2L predictions over a mitigated emitter in Kazakhstan, Middle Caspian Basin}. Top: a timeseries of validated emissions with quantification. Bottom: Examples of MARS-S2L predictions compared to expert-annotated ground-truth.}
    \label{fig:mitigation_kazakhstan}
\end{figure}

\begin{figure}
    \centering
    \includegraphics[width=0.8\linewidth]{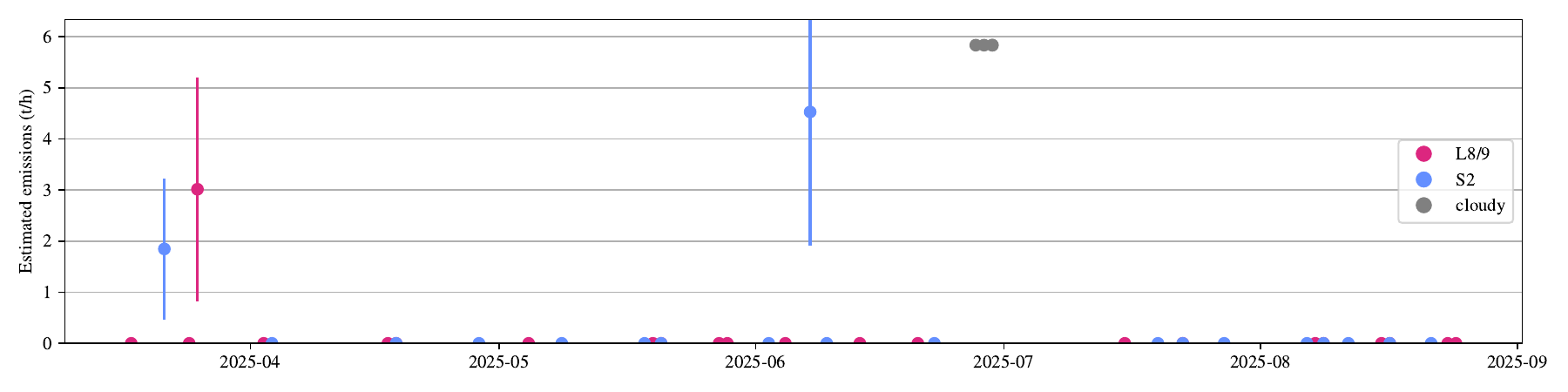}\\
    \includegraphics[width=0.8\linewidth]{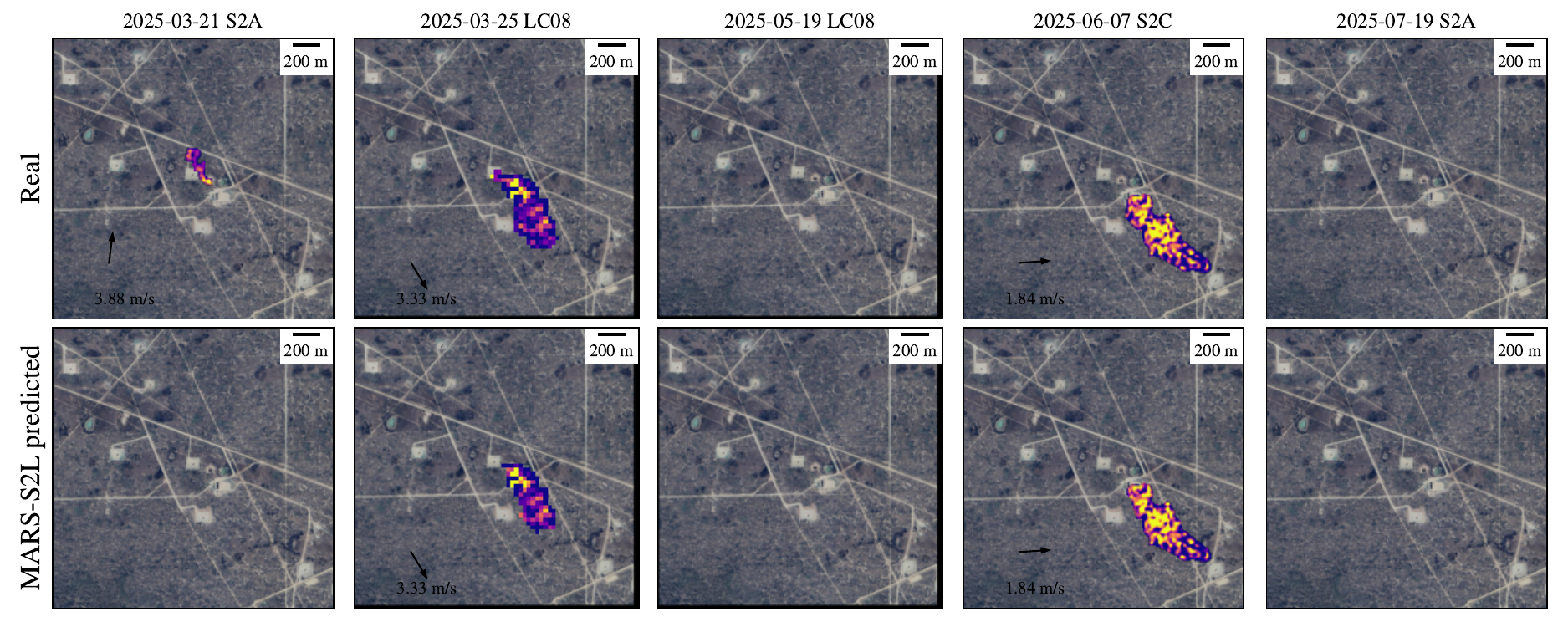}\\
    \includegraphics[width=0.2\linewidth]{figsmitigation/pretty_colorbar_horizontal_2000.pdf}
    \caption{\textbf{MARS-S2L predictions over a mitigated emitter in Argentina, Meseta Espinosa IV}. Top: a timeseries of validated emissions with quantification. Bottom: Examples of MARS-S2L predictions compared to expert-annotated ground-truth.}
    \label{fig:mitigation_argentina}
\end{figure}

\begin{figure}
    \centering
    \includegraphics[width=0.8\linewidth]{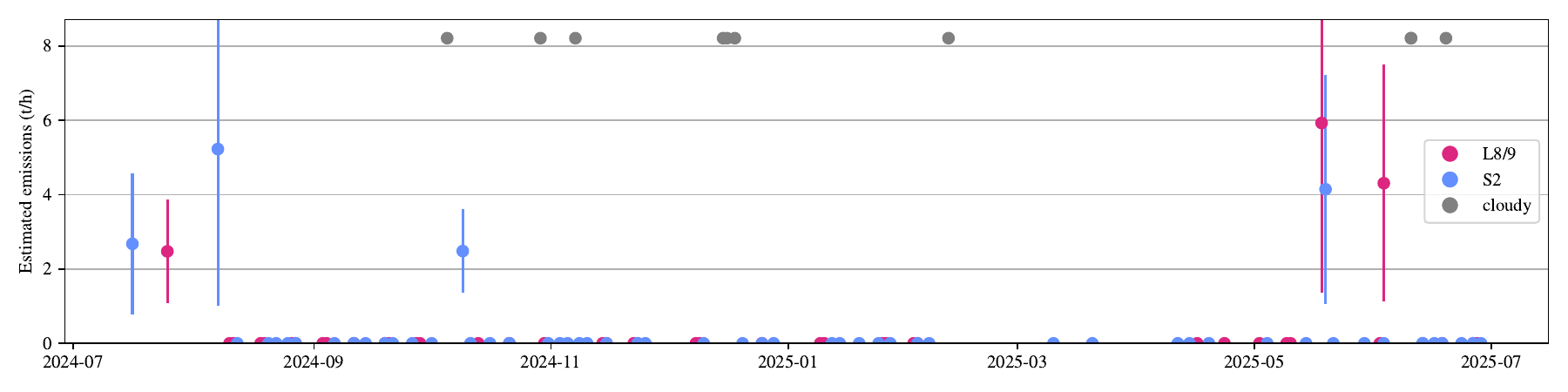}\\
    \includegraphics[width=0.8\linewidth]{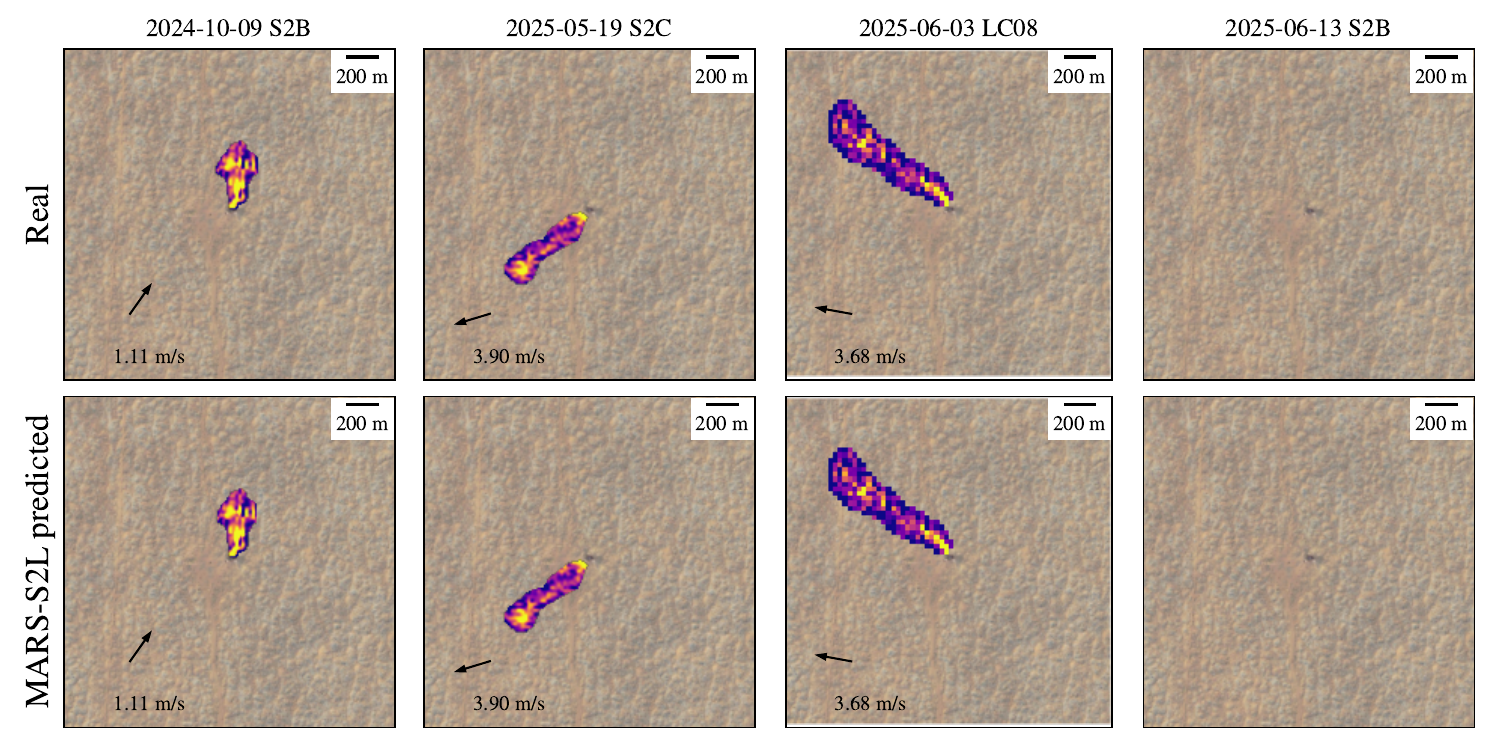}\\
    \includegraphics[width=0.2\linewidth]{figsmitigation/pretty_colorbar_horizontal_2000.pdf}
    \caption{\textbf{MARS-S2L predictions over a mitigated emitter in Turkmenistan, Northern Balguyy}. Top: a timeseries of validated emissions with quantification. Bottom: Examples of MARS-S2L predictions compared to expert-annotated ground-truth.}
    \label{fig:mitigation_turkmenistan}
\end{figure}

In this subsection we show operational model detections over the additional 5 locations with confirmed mitigation actions in Libya (Fig.~\ref{fig:mitigation_libya}), Yemen (Fig.~\ref{fig:mitigation_yemen}), Kazakhstan (Fig.~\ref{fig:mitigation_kazakhstan}), Argentina (Fig.~\ref{fig:mitigation_argentina}) and Turkmenistan (Fig.~\ref{fig:mitigation_turkmenistan}).

\subsubsection*{Model prediction examples}
Example timeseries of model predictions are shown in Figures S18-S25 for emitters in Thailand, Bahrain, Venezuela, Libya, the US, Algeria, Turkmenistan and Yemen. These figures showcase model performance across a range of background and emitter types in a diverse range of regions globally. 

\newpage

\begin{figure}
    \centering
    \includegraphics[height=.35\paperheight]{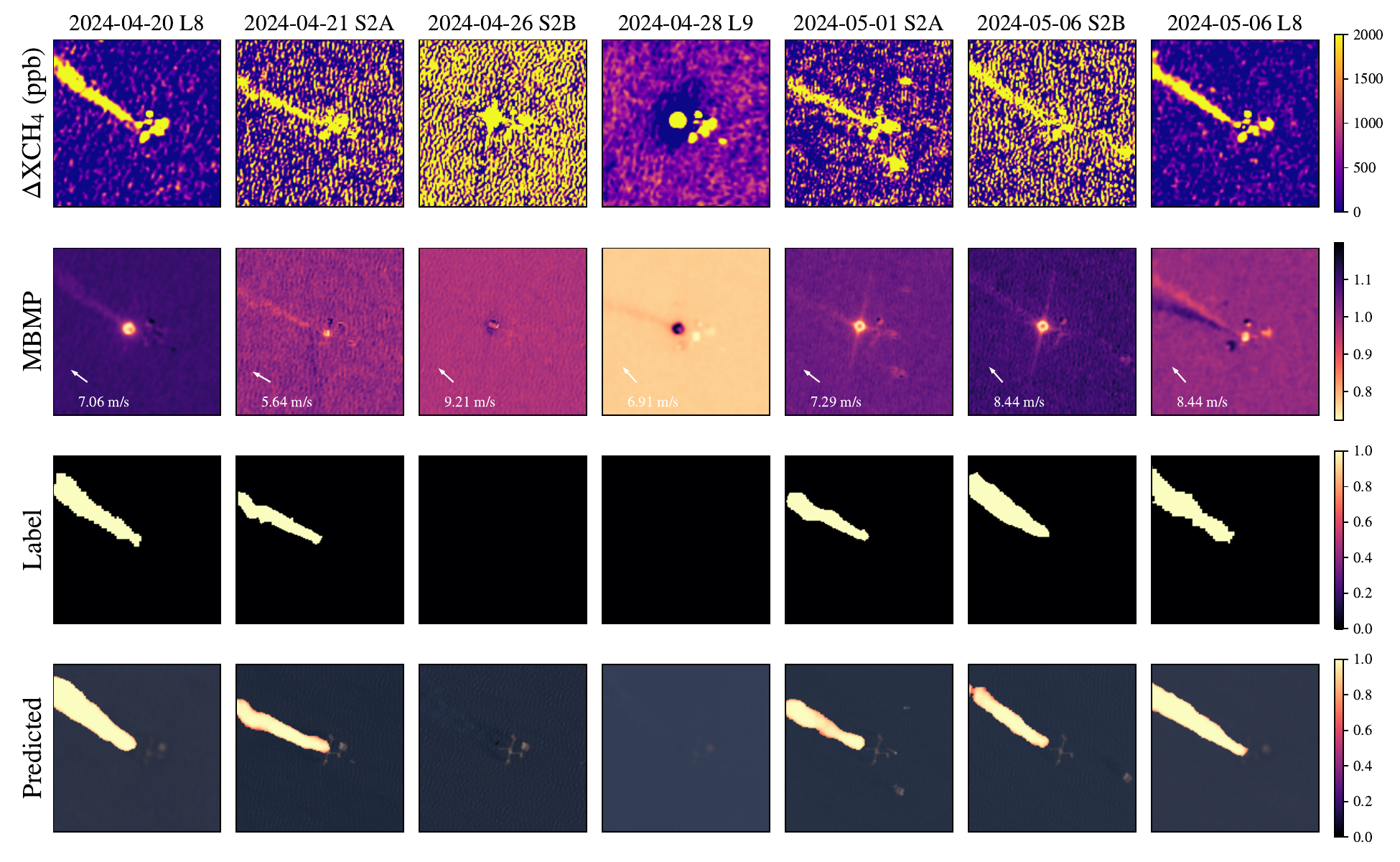}\\
    \caption{MARS-S2L predictions for an emitter in the gulf of Mexico.}
    \label{fig:thai}
\end{figure}

\begin{figure}
    \centering
    \includegraphics[height=.35\paperheight]{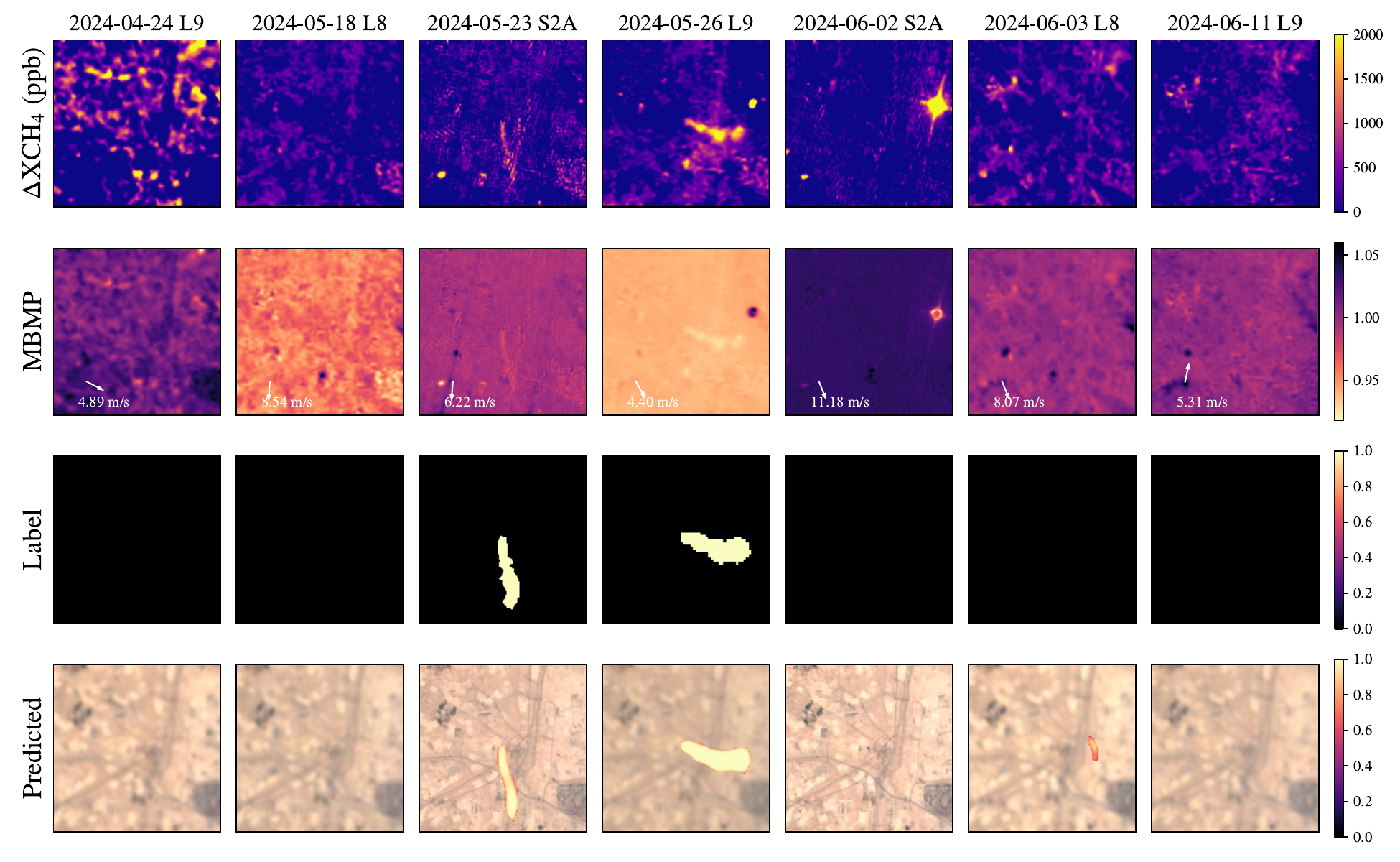}\\
    \caption{MARS-S2L predictions for an emitter in Bahrain.}
    \label{fig:bah}
\end{figure}

\begin{figure}
    \centering
    \includegraphics[height=.35\paperheight]{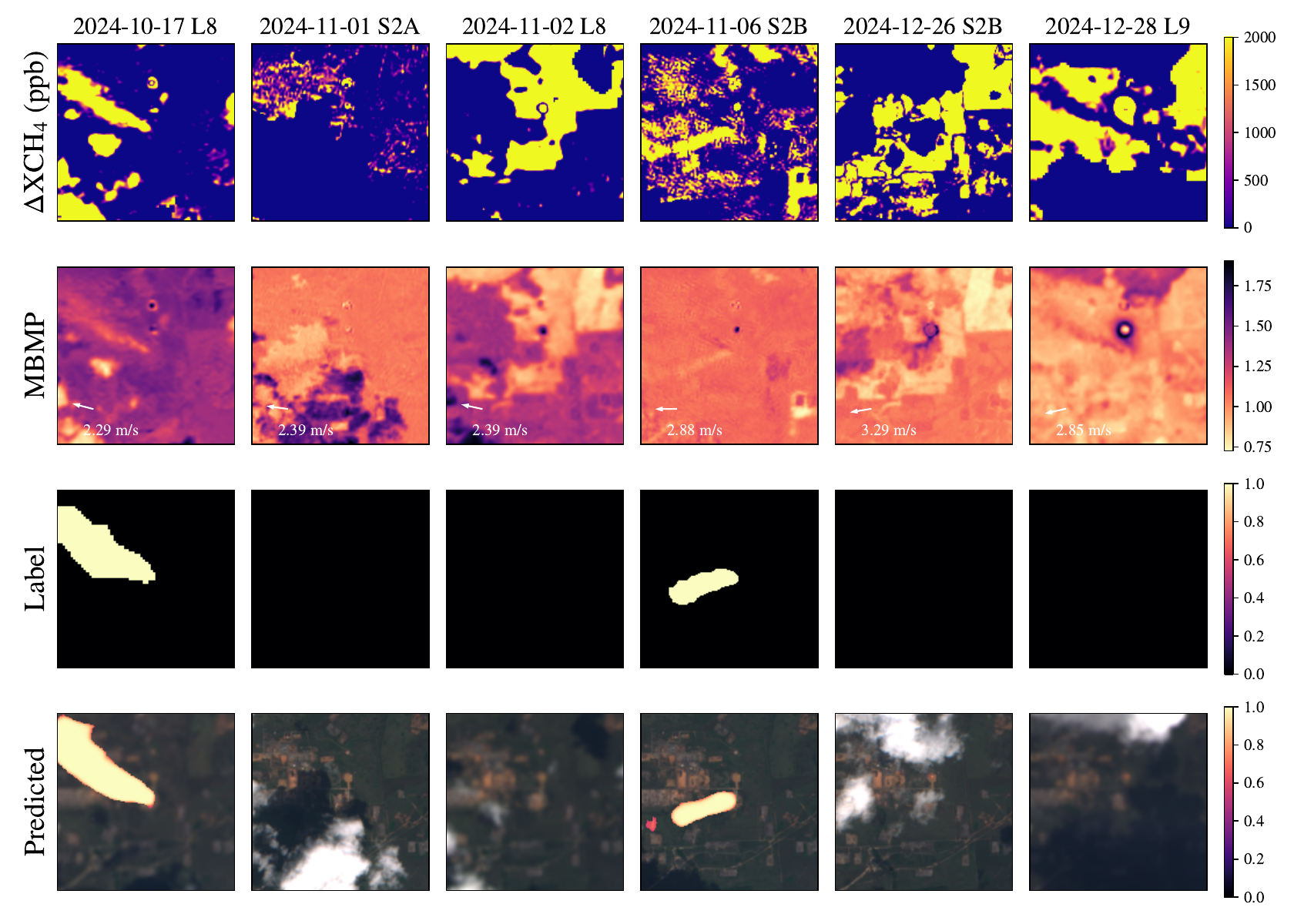}\\
    \caption{MARS-S2L predictions for an emitter in Venezuela.}
    \label{fig:ven}
\end{figure}

\begin{figure}
    \centering
    \includegraphics[height=.35\paperheight]{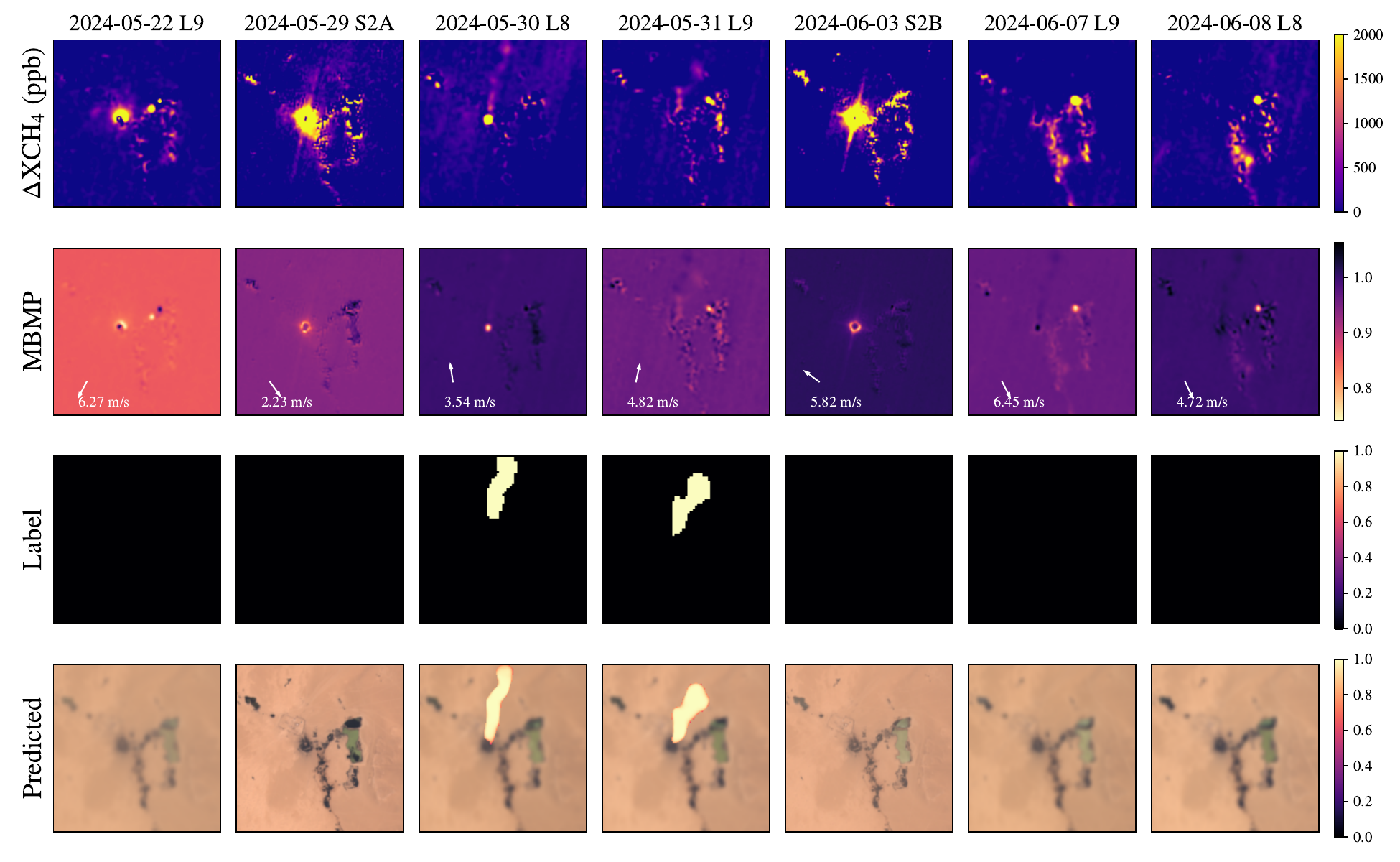}\\
    \caption{MARS-S2L predictions for an emitter in Libya.}
    \label{fig:lib}
\end{figure}

\begin{figure}
    \centering
    \includegraphics[height=.35\paperheight]{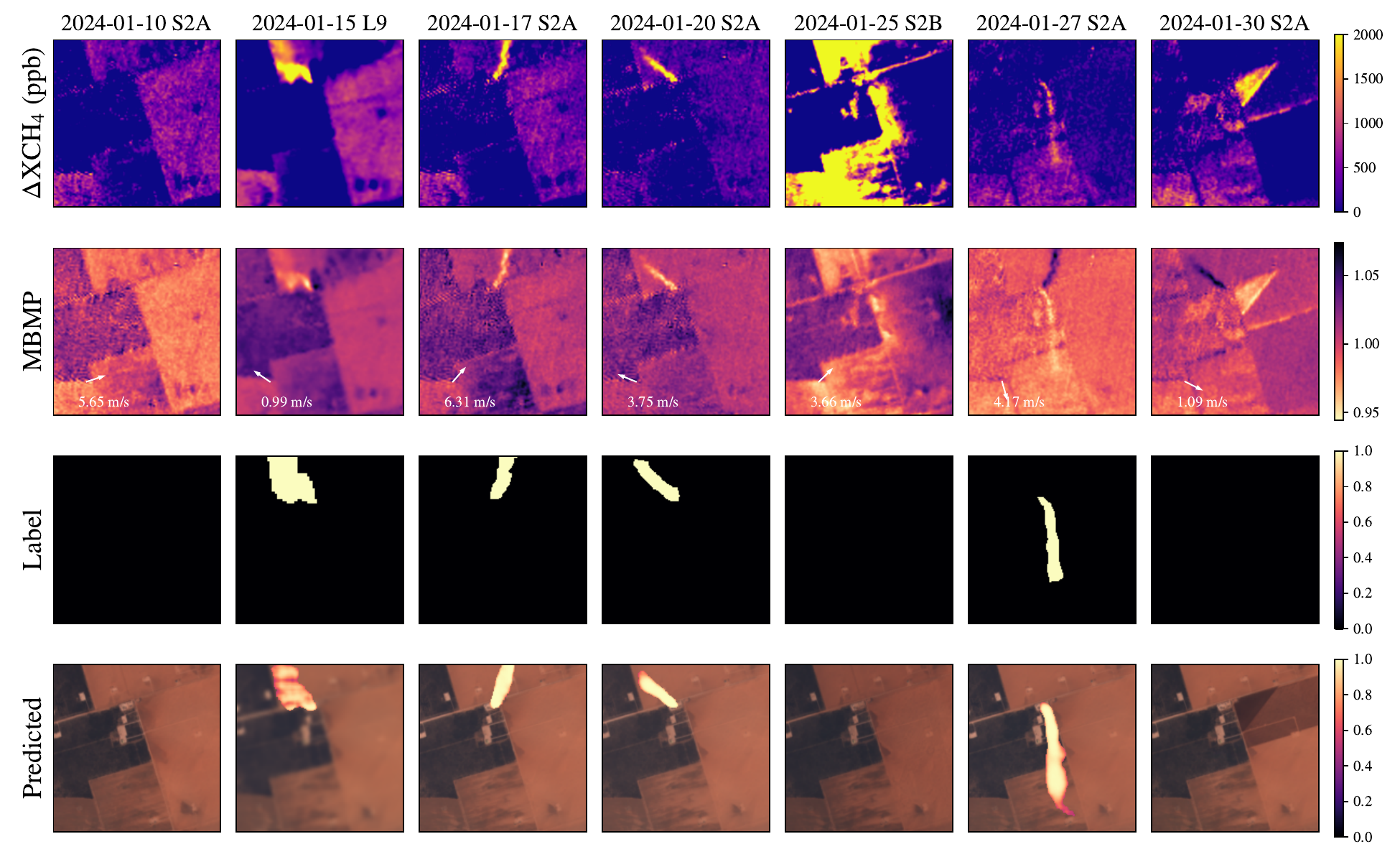}\\
    \caption{MARS-S2L predictions for an emitter in the US.}
    \label{fig:us}
\end{figure}

\begin{figure}
    \centering
    \includegraphics[height=.35\paperheight]{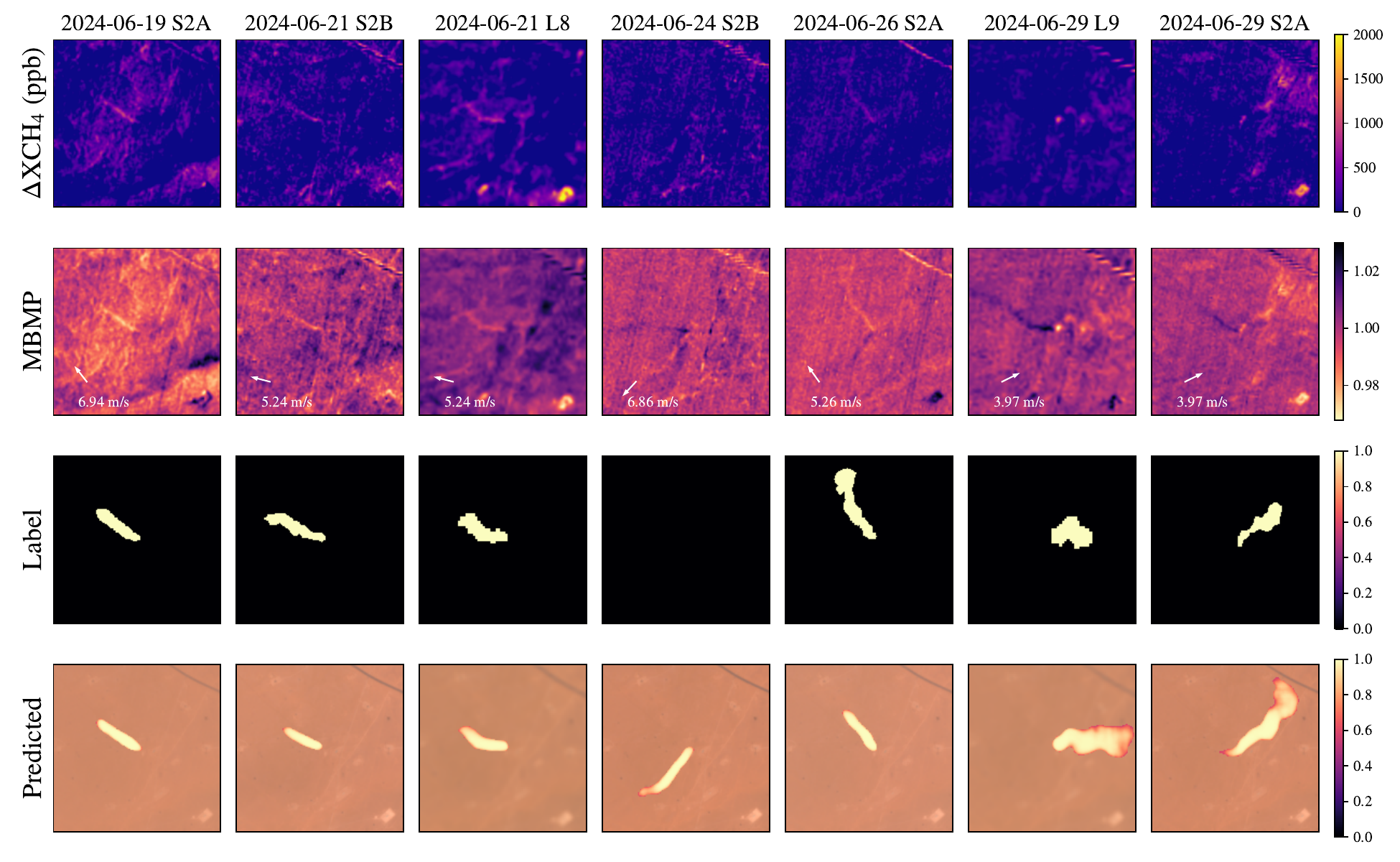}\\
    \caption{MARS-S2L predictions for an emitter in Algeria.}
    \label{fig:dza}
\end{figure}

\begin{figure}
    \centering
    \includegraphics[height=.35\paperheight]{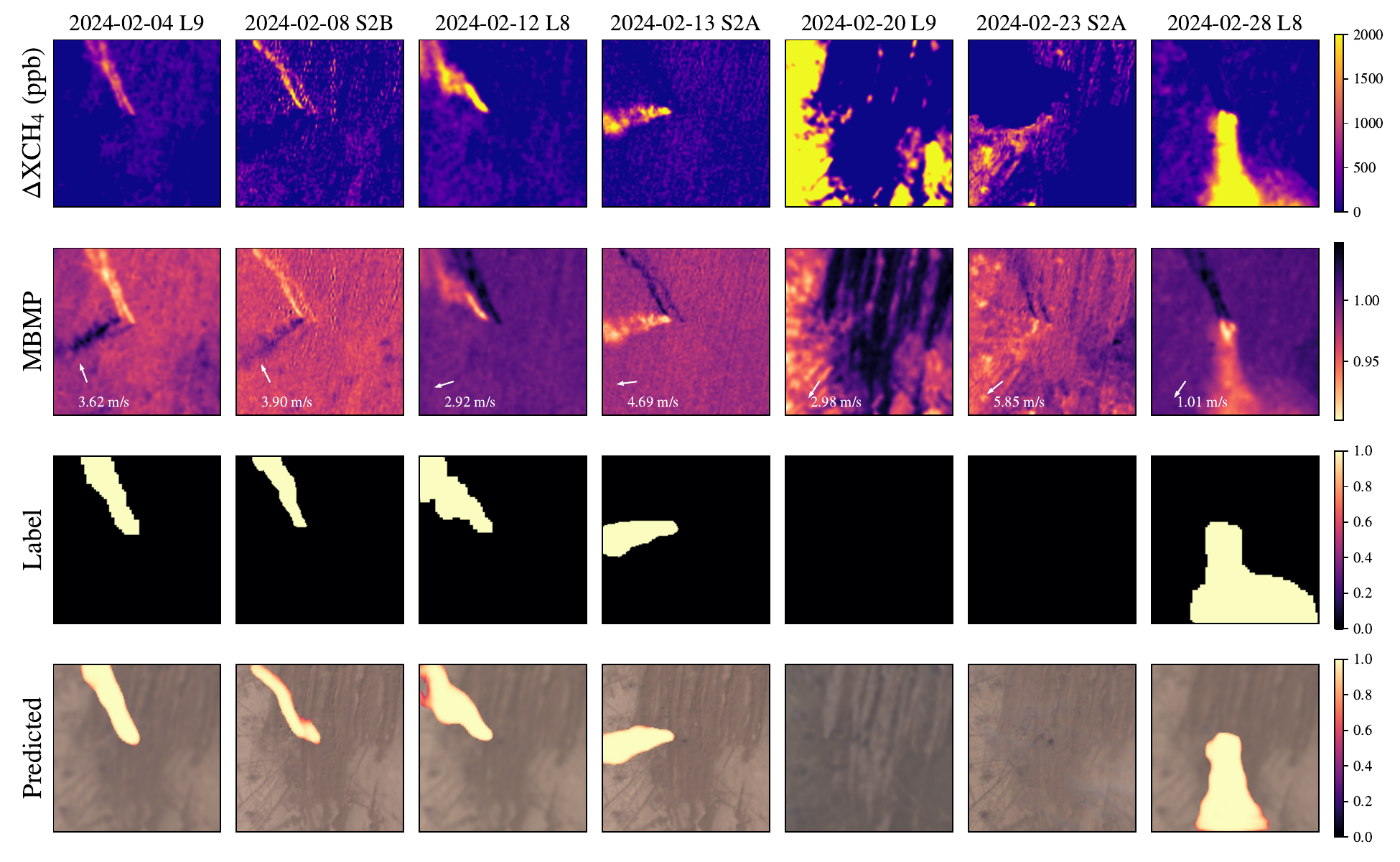}\\
    \caption{MARS-S2L predictions for an emitter in Turkmenistan.}
    \label{fig:turk}
\end{figure}

\begin{figure}
    \centering
    \includegraphics[height=.35\paperheight]{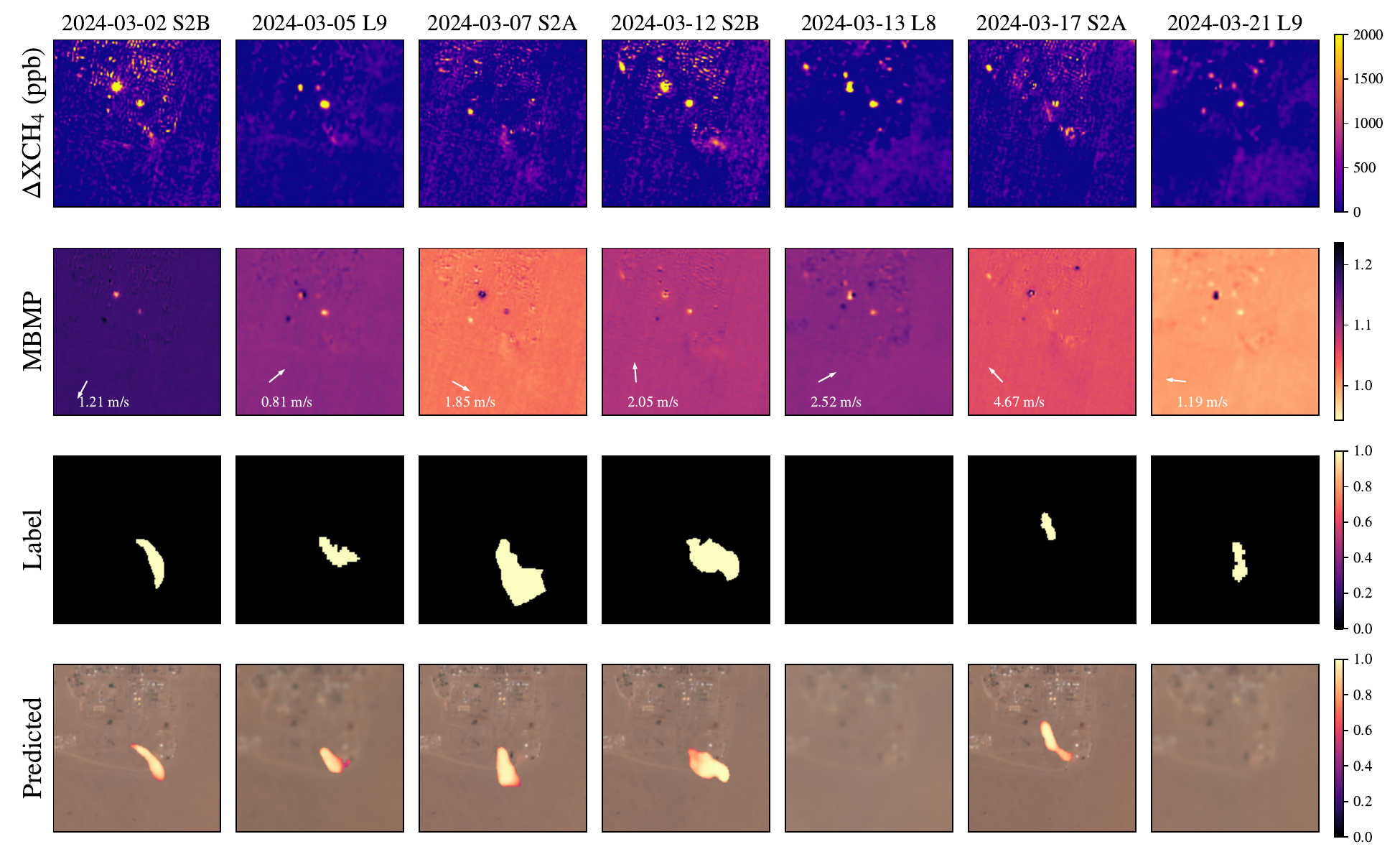}\\
    \caption{MARS-S2L predictions for an emitter in Yemen.}
    \label{fig:yem}
\end{figure}

\subsubsection*{The MARS-S2L dataset}

\begin{figure}
    \centering
    \includegraphics[height=.7\paperheight]{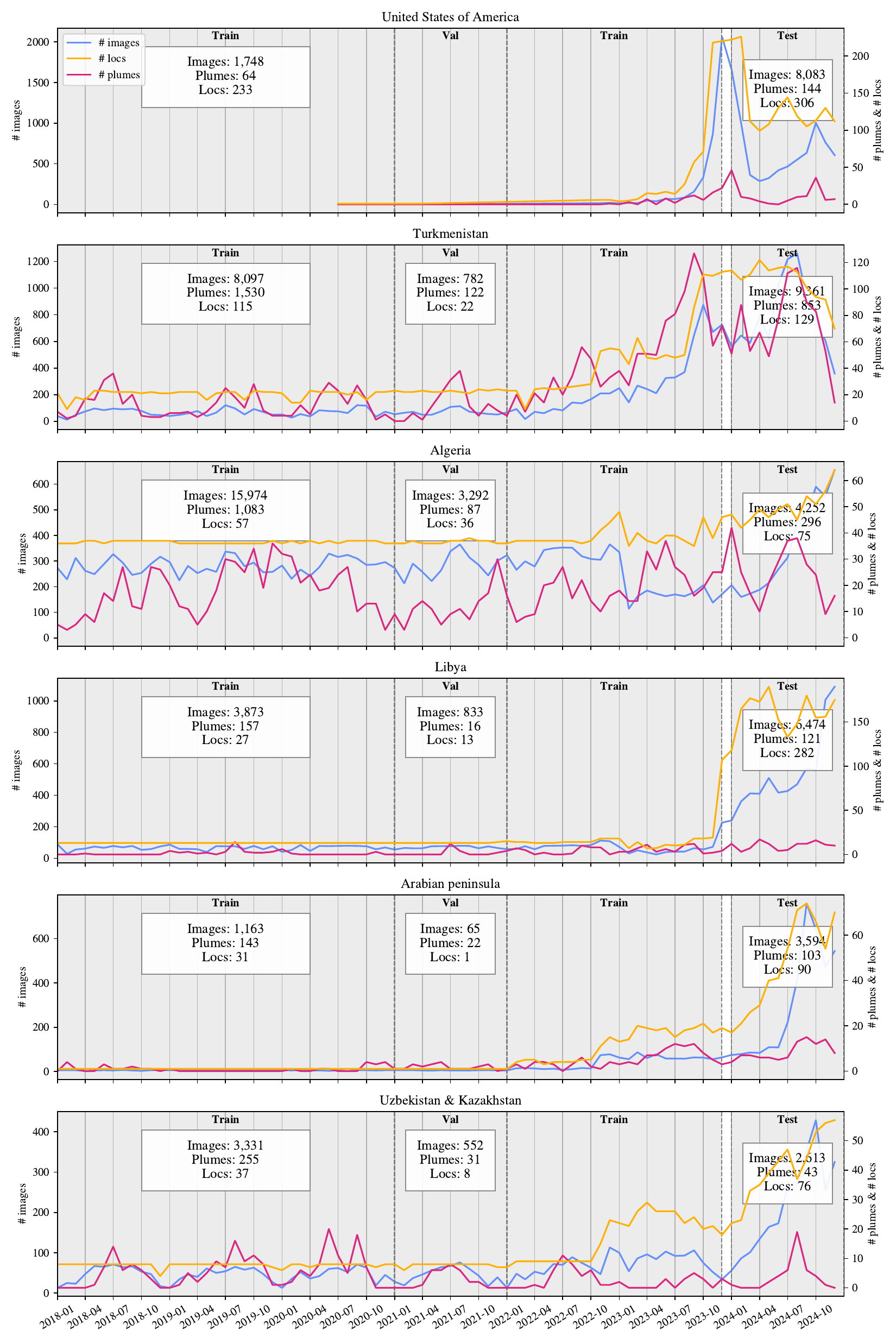}\\
    \caption{Monthly number of images, plumes and sources on first six case studies.}
    \label{fig:nimagesplumesbycountrymars2}
\end{figure}

\begin{figure}
    \centering
    \includegraphics[height=.7\paperheight]{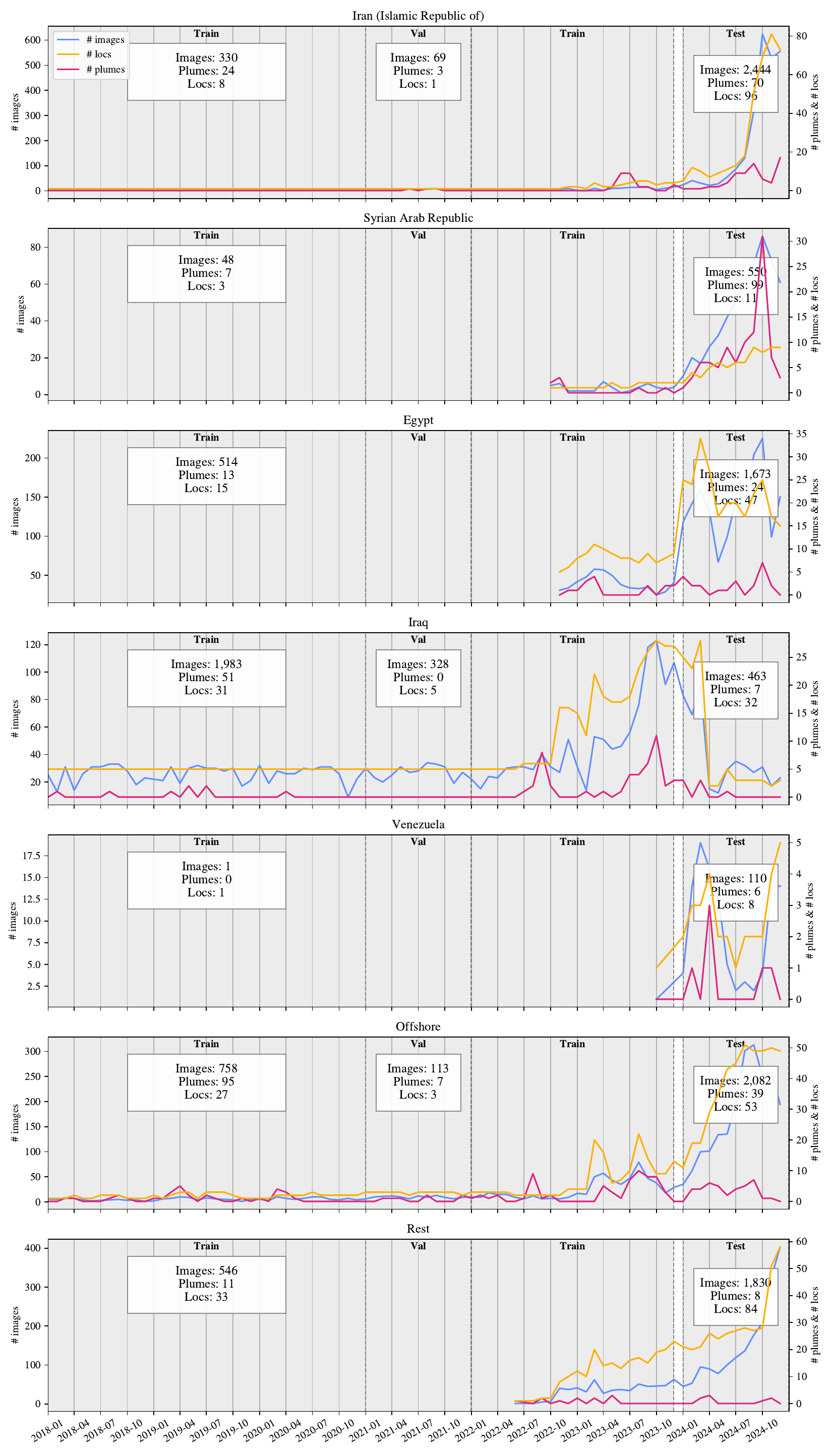}\\
    \caption{Monthly number of images, plumes and sources on rest of case studies studies.}
    \label{fig:nimagesplumesbycountryrest}
\end{figure}

This section presents supplementary information of the MARS-S2L dataset. In total, the dataset compiled as part of this work comprises images of 1,315 emitters in 38 countries at the time of this analysis. Figures \ref{fig:nimagesplumesbycountrymars2} and \ref{fig:nimagesplumesbycountryrest} show the number of images, locations and plumes analysed over each country per month for 2023 and 2024 including the total number of images and locations used for training, validation and testing.

\newpage

\clearpage

\subsubsection*{Extra tables}
\label{sec:extra_figures}

In this section we provide supplementary information with the statistics derived in the main text. Tables~\ref{tab:notified_plumes} and~\ref{tab:notified_plumes_by_month} show the number of detected and notified plumes by country and by month, respectively, by the Methane Alert and Response System. Table~\ref{tab:case_study_metrics} shows the metrics of the different models by case-study area. Table~\ref{tab:model_metrics_thresholds} shows the overall metrics for different probability thresholds.

\newpage
 
\begin{table}[]
    \centering
\begin{tabular}{lrr}
\toprule
Country & Sites & Plumes notified \\
\midrule
Turkmenistan & 110 & 1374 \\
Algeria & 98 & 339 \\
Iran (Islamic Republic of) & 40 & 277 \\
Syrian Arab Republic & 19 & 227 \\
United States of America & 65 & 144 \\
Libya & 35 & 118 \\
Yemen & 8 & 60 \\
Egypt & 15 & 50 \\
Bahrain & 6 & 35 \\
Mexico & 2 & 35 \\
Kazakhstan & 12 & 27 \\
Kuwait & 5 & 25 \\
Uzbekistan & 16 & 22 \\
Russian Federation & 6 & 7 \\
Saudi Arabia & 3 & 6 \\
Jordan & 2 & 6 \\
Malaysia & 1 & 5 \\
China & 3 & 4 \\
Venezuela & 2 & 4 \\
Iraq & 4 & 4 \\
Argentina & 2 & 2 \\
Australia & 1 & 2 \\
Azerbaijan & 1 & 1 \\
Oman & 1 & 1 \\
Italy & 1 & 1 \\
\bottomrule
\end{tabular}
    \caption{\textbf{Formal government notifications issued after a MARS-S2L plume detection}. Number of unique sites and plumes by country detected by the MARS-S2L model formally notified to country government representatives between \NotificationStartDate{} to \NotificationEndDate{}. In total \NotificationPlumes{} plumes from \NotificationSources{} different sources across \NotificationCountries{} different countries.}
    \label{tab:notified_plumes}
\end{table}

\begin{table}
    \centering
\begin{tabular}{lrr}
\toprule
Year-Month & Sites & Plumes notified \\
\midrule
2024-01 & 10 & 12 \\
2024-03 & 8 & 8 \\
2024-04 & 4 & 4 \\
2024-05 & 6 & 6 \\
2024-06 & 40 & 100 \\
2024-07 & 52 & 139 \\
2024-08 & 54 & 151 \\
2024-09 & 53 & 136 \\
2024-10 & 63 & 146 \\
2024-11 & 35 & 71 \\
2024-12 & 30 & 47 \\
2025-01 & 34 & 71 \\
2025-02 & 28 & 44 \\
2025-03 & 35 & 79 \\
2025-04 & 33 & 62 \\
2025-05 & 60 & 134 \\
2025-06 & 69 & 207 \\
2025-07 & 67 & 198 \\
2025-08 & 67 & 226 \\
2025-09 & 72 & 196 \\
2025-10 & 97 & 262 \\
2025-11 & 84 & 256 \\
2025-12 & 43 & 73 \\
2026-01 & 60 & 105 \\
2026-02 & 35 & 43 \\
\bottomrule
\end{tabular}
\caption{\textbf{Formal government notifications issued after a MARS-S2L plume detection}. Number of unique sites and plumes by month detected by the MARS-S2L model formally notified to country government representatives between \NotificationStartDate{} to \NotificationEndDate{}. In total \NotificationPlumes{} plumes from \NotificationSources{} different sources across \NotificationCountries{} different countries.}
\label{tab:notified_plumes_by_month}
\end{table}

\begin{table}[h]
    \centering
    \footnotesize
\begin{tabular}{llrrrr}
\toprule
Case study & Model & AP (\%) & Precision (\%) & Recall (\%) & FPR (\%) \\
\midrule
\multirow{3}{*}{Turkmenistan}  & MARS-S2L & 75.52 & 50.00 & 80.66 & 8.09 \\
 & CH4Net & 40.86 & 26.32 & 63.54 & 17.83 \\
 & MBMP & 10.75 & 11.65 & 81.83 & 62.20 \\
\hline
\multirow{3}{*}{United States of America} & MARS-S2L & 74.67 & 26.91 & 85.42 & 4.21 \\
 & CH4Net & 3.54 & 6.67 & 8.33 & 2.12 \\
 & MBMP & 2.11 & 2.07 & 100.00 & 85.72 \\
\hline
\multirow{3}{*}{Libya} & MARS-S2L & 31.24 & 12.56 & 68.60 & 9.10 \\
 & CH4Net & 3.59 & 3.38 & 20.66 & 11.24 \\
 & MBMP & 3.79 & 2.95 & 78.51 & 49.16 \\
\hline
\multirow{3}{*}{Algeria} & MARS-S2L & 72.98 & 49.58 & 79.05 & 6.02 \\
 & CH4Net & 28.84 & 21.41 & 56.42 & 15.50 \\
 & MBMP & 8.55 & 9.50 & 71.62 & 51.04 \\
\hline
\multirow{3}{*}{Arabian peninsula} & MARS-S2L & 53.19 & 23.75 & 78.64 & 7.45 \\
 & CH4Net & 8.69 & 8.27 & 42.72 & 13.98 \\
 & MBMP & 2.52 & 2.80 & 65.05 & 66.66 \\
\hline
\multirow{3}{*}{Uzbekistan \& Kazakhstan} & MARS-S2L & 36.66 & 13.33 & 74.42 & 8.09 \\
 & CH4Net & 2.50 & 0.79 & 4.65 & 9.81 \\
 & MBMP & 1.06 & 1.27 & 58.14 & 75.60 \\
\hline
\multirow{3}{*}{Iran} & MARS-S2L & 64.26 & 24.44 & 78.57 & 7.16 \\
 & CH4Net & 4.90 & 4.52 & 14.29 & 8.89 \\
 & MBMP & 2.65 & 3.14 & 97.14 & 88.46 \\
\hline
\multirow{3}{*}{Offshore} & MARS-S2L & 53.46 & 18.99 & 87.18 & 7.10 \\
 & CH4Net & 59.43 & 22.36 & 92.31 & 6.12 \\
 & MBMP & 4.88 & 1.87 & 100.00 & 100.00 \\
\hline
\multirow{3}{*}{Egypt} & MARS-S2L & 33.99 & 11.69 & 75.00 & 8.25 \\
 & CH4Net & 1.95 & 1.82 & 4.17 & 3.27 \\
 & MBMP & 1.14 & 0.98 & 37.50 & 55.12 \\
\hline
\multirow{3}{*}{Syria} & MARS-S2L & 65.16 & 51.06 & 72.73 & 15.30 \\
 & CH4Net & 17.72 & 24.44 & 11.11 & 7.54 \\
 & MBMP & 14.19 & 15.47 & 54.55 & 65.41 \\
\hline
\multirow{3}{*}{Iraq} & MARS-S2L & 54.47 & 12.50 & 71.43 & 7.68 \\
 & CH4Net & 2.94 & 0.00 & 0.00 & 10.31 \\
 & MBMP & 1.40 & 1.67 & 85.71 & 77.41 \\
\hline
\multirow{3}{*}{Venezuela} & MARS-S2L & 84.90 & 31.25 & 83.33 & 10.58 \\
 & CH4Net & 21.00 & 0.00 & 0.00 & 0.00 \\
 & MBMP & 9.17 & 5.45 & 100.00 & 100.00 \\
\bottomrule
\end{tabular}
    \caption{Model performance across the different geographical case-studies.}
    \label{tab:case_study_metrics}
\end{table}

\begin{table}[]
    \centering
    \begin{tabular}{lrrrr}
\toprule
Model & Threshold & Precision (\%) & Recall (\%) & FPR (\%) \\
\midrule
MARS-S2L & 0.50 & 30.78 & 79.04 & 7.73 \\
CH4Net & 0.50 & 16.44 & 46.94 & 10.37 \\
MBMP & 0.99 & 4.74 & 78.87 & 68.88 \\
\hline
MARS-S2L & 0.90 & 51.99 & 69.22 & 2.78 \\
CH4Net & 0.90 & 29.14 & 27.74 & 2.93 \\
MBMP & 0.90 & 2.81 & 9.10 & 13.69 \\
\hline
MARS-S2L & 0.98 & 68.61 & 58.36 & 1.16 \\
CH4Net & 0.98 & 41.15 & 16.55 & 1.03 \\
MBMP & 0.85 & 2.66 & 4.63 & 7.37 \\
\bottomrule
\end{tabular}
    \caption{Model performance metrics for different probability thresholds. Results on previous tables use the 0.5 threshold for the MARS-S2L and CH4Net models and 0.99 for MBMP.}
    \label{tab:model_metrics_thresholds}
\end{table}

\end{document}